%% file: 0_main.tex
\pdfoutput=1

\documentclass[11pt]{article}

\usepackage[]{acl}

\usepackage{times}
\usepackage{latexsym}
\usepackage{graphicx}
\usepackage{comment}
\usepackage{lipsum}
\usepackage{footnote}
\usepackage{hyperref}
\usepackage{subcaption}
\usepackage{booktabs}
\usepackage{longtable}
\usepackage{pdflscape}
\usepackage{adjustbox}
\usepackage{caption}

\usepackage{enumitem}
\setlist{nolistsep,leftmargin=*}

\usepackage{multirow}
\usepackage{longtable}
\usepackage{graphicx}
\usepackage{adjustbox}

\usepackage[T1]{fontenc}

\usepackage[utf8]{inputenc}

\usepackage[draft,textsize=footnotesize,textwidth=15mm]{todonotes}

\usepackage{microtype}

\title{

IMAGINATOR: Pre-Trained Image+Text Joint Embeddings using Word-Level Grounding of Images }

\author{
Varuna Krishna\textsuperscript{1} \quad 
S Suryavardan \textsuperscript{1} \quad
Shreyash Mishra\textsuperscript{1}\quad \\
\bf Sathyanarayanan Ramamoorthy\textsuperscript{2}  \quad
\bf Parth Patwa\textsuperscript{3} \quad 
\bf Megha Chakraborty\textsuperscript{4} \quad \\
\bf Aman Chadha\dag\textsuperscript{5,6}  \quad
\bf Amitava Das\textsuperscript{4} \quad 
\bf Amit Sheth\textsuperscript{4} \quad \\
\textsuperscript{1}IIIT Sri City, India \quad 
\textsuperscript{2}CMU, USA \quad
\textsuperscript{3}UCLA, USA \\
\textsuperscript{4}University of South Carolina, USA \quad
\textsuperscript{5}Amazon AI, USA \quad
\textsuperscript{6}Stanford University, USA\\
\tt  varunakrishna.k19@iiits.in \quad
\tt amitava@mailbox.sc.edu
}

\begin{document}

\maketitle

\renewcommand{\thefootnote}{\fnsymbol{footnote}}
\footnotetext[2]{Work does not relate to position at Amazon.}
\renewcommand*{\thefootnote}{\arabic{footnote}}
\setcounter{footnote}{0}

\begin{abstract}
Word embeddings, i.e., semantically meaningful vector representation of words, are largely influenced by the distributional hypothesis \textit{"You shall know a word by the company it keeps"} \cite{distributional}, whereas modern prediction-based neural network embeddings rely on design choices and hyperparameter optimization. Word embeddings like Word2Vec, GloVe  etc. well capture the contextuality and real-world analogies but contemporary convolution-based image embeddings such as VGGNet, AlexNet, etc. do not capture contextual knowledge. The popular \textit{\textit{king-queen}} analogy does not hold true for most commonly used vision embeddings.

In this paper, we introduce a pre-trained joint embedding (JE), named IMAGINATOR, trained on 21K distinct image objects level from 1M image+text pairs. JE is a way to encode multimodal data into a vector space where the text modality serves as the grounding key, which the complementary modality (in this case, the image) is anchored with. IMAGINATOR encapsulates three individual representations: \textit{(i) object-object co-location, (ii) word-object co-location, and (iii) word-object correlation}. These three ways capture complementary aspects of the two modalities which are further combined to obtain the final JEs.

Generated JEs are intrinsically evaluated to assess how well they capture the contextuality and real-world analogies. We also evaluate pre-trained IMAGINATOR JEs on three downstream tasks: (i) image captioning, (ii) Image2Tweet, and (iii) text-based image retrieval. IMAGINATOR establishes a new standard on the aforementioned downstream tasks by outperforming the current SoTA on all the selected tasks. IMAGINATOR will be made publicly available. The codes are available at \url{https://github.com/varunakk/IMAGINATOR}
\end{abstract}

\input{introduction}

\input{1_introduction}
\input{2_hypothesis}
\input{3_lessons_from_NLP}
\input{4_learning_JE}
\input{5_intrinsic}
\input{6_extrinsic}

\input{8_conclusion}

\bibliography{anthology}
\bibliographystyle{acl_natbib}


\newpage
\onecolumn
\section*{Frequently Asked Questions (FAQs)}

\begin{enumerate}
    \item \textbf{Doesn't averaging individual object embeddings (or even word embeddings) result in a noisy object embedding?}
    \begin{description}
     \item \textbf{Ans.} Yes, averaging individual embeddings is a limitation of this work and a future avenue of exploration. On the other hand, concatenation is computationally expensive. However, empirically, we found that averaging gave better results than concatenation. We would like to explore autoencoding and contrastive learning in the future as mitigation methods. 
     \end{description}
    \bigskip
    \item \textbf{Why does orthogonal projection work better than CCA-based methods?}\begin{description}
     \item \textbf{Ans.} Orthogonal projection is a discriminative method that attempts to find out the discriminative projection for two vector spaces aligned to a unified dimension. On the other hand, CCA tries to learn relations among two vector spaces. While orthogonal projection offers competitive performance with a limited number of classes, CCA is undoubtedly more powerful when the number of classes is higher. In our case, since we only have 21K objects, orthogonal projection yielded better results.
     \end{description}
\bigskip
     
    \item \textbf{Instead of directly learning a caption generation model based on the learned joint embedding, this paper projects VGG-19 embeddings orthogonally in the learned joint embedding space, using it to find the $k$ nearest objects in the vector space, and then passes these objects through T5 for caption generation. What is the motivation behind this approach?}
    \begin{description}
     \item \textbf{Ans.} Image object detection is a separate task altogether, and we are not trying to solve that problem here. Given an image, we first get its VGG-19 embedding and then project it to IMAGINATOR space since VGG-19 and IMAGINATOR have disparate embedding spaces and need to be aligned. A by-product of this approach is that it also helps affirm that IMAGINATOR performs well, otherwise it might raise doubts that the performance gain is happening due to T5 efficiency rather than IMAGINATOR. Lastly, we would like to draw the attention of readers that the proposed captioning architecture is very simple and still outperforms SoTA. 
    \end{description}
\bigskip
    
    \item \textbf{Did you consider experimenting with ResNet or Fast-RCNN?}
    \begin{description}
     \item \textbf{Ans.} We performed experiments using ResNet, but the results were poor. One plausible explanation is the fact that higher embedding dimensions lead to a performance drop.
     \end{description}
\bigskip
     
    \item \textbf{Why was the Detic the baseline architecture of choice for IMAGINATOR?}
    \begin{description}
     \item \textbf{Ans.} The presumption of this work is to leverage the legacy of the NLP-centric count-based vectorification methods for joint modality. Therefore, maximizing the number of objects will give us a denser matrix to calculate the so-called co-location. In the future, we plan to seek methods that can detect more than 21K objects, and strongly believe that will have a positive effect on the learned joint embedding space.   
     \end{description}
\end{enumerate}

\input{9_appendix}


\end{document}

%% file: introduction.tex
\section{Joint Modality and Contextuality}
\begin{figure}[htp]
    \centering
    \includegraphics[width=0.47\textwidth]{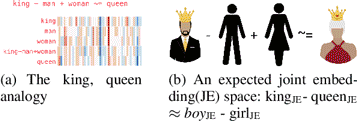}
    \caption{CNN-based image embeddings are unable to capture contextuality like existing word embeddings. The \textit{king-queen} vs. \textit{man-woman} analogy has been popularized by \cite{mikolov-etal-2013-linguistic}, whereas drawing a similar analogy in image vector space is rather difficult. We argue joint embedding is the alternative.}
    \label{fig:img1}
\end{figure}

Word embeddings are learned representations such that words with similar meanings are represented similarly. Distribution-based compositional word embeddings like Word2vec \cite{Mikolov2013EfficientEO} and GloVe \cite{pennington-etal-2014-glove} are popular in modern NLP. These are used to extract the notion of relatedness across different words, and capture the overall semantic meaning of a text. Consider the \textit{king-queen} \cite{mikolov-etal-2013-linguistic} word vector analogy (figure \ref{fig:img1}), which shows how good these word embeddings are at capturing syntactic and semantic regularities in language. 

The notion of contextual similarity (i.e., words occurring together) is used in learning the representations, because of which vector arithmetic like {\tt King - Man + Woman = Queen} are possible. See figure ~\ref{fig:img1}~\cite{mikolov-etal-2013-linguistic}. Deriving an analogous representation using images is a challenging task since the concept of relatedness among images is not well-defined. Motivated by this argument, we propose creating joint embeddings (JEs) that can represent real-world analogies, which can aid in solving several multimodal tasks owing to their distributional semantics.
 
%% file: 1_introduction.tex
\section{Contemporary Joint Embedding Methods }\label{sec:related_work}

Canonical Correlation Analysis (CCA) based methods use similarities to project two inputs onto a vector space. CLIP \cite{radford2021learning} utilizes contrastive pre-training and encodes aligned image and text embeddings with the help of text and visual modality encoders. Stanford’s Joint Embedding \cite{Kolluru2019ANA} uses VGG-19 \cite{simonyan2014very} and GLoVe \cite{pennington-etal-2014-glove} to generate the image and text encodings using a triplet loss. \citet{chen2020uniter} proposed UNITER, trained on a large dataset, which uses an image and text encoder and a transformer to generate the final embeddings. \citet{jia2021scaling} use a noisy dataset of 1 billion (image, alt-text) pairs and propose a dual architecture for aligning and generating the visual and textual embeddings. This architecture uses contrastive loss for learning. \citet{tan2019lxmert} proposed a framework to create a relation between visual and language modalities. This architecture consists of three encoders, one object relation encoder, a language encoder and a cross-modal encoder. Compared to the aforementioned prior works, illustrated in appendix figure \ref{fig:img2}, the unique differentiating factor with IMAGINATOR is that we focus on the word-level grounding \cite{Gunti-Ramamoorthy-Patwa-Das-2022} of images while prior works perform embedding generation at the sentence level. Our belief is that this will help us learn rich relational features, i.e., features that are rich encapsulations of words and the corresponding objects they represent via images. 

%% file: 2_hypothesis.tex
\section{IMAGINATOR - Learning Joint Embeddings}
Off-the-shelf word embeddings like Word2vec \cite{Mikolov2013EfficientEO}, GloVe \cite{pennington-etal-2014-glove} and the embeddings generated by BERT \cite{devlin2018bert}, GPT \cite{radford2021learning} are used for tacking several downstream NLP tasks. The motivation behind creating IMAGINATOR is to have similar pre-trained embeddings for vision-language tasks. Researchers can download pre-trained JEs and utilize them for any vision-language task they have in hand. Existing techniques have only explored JEs from the sentence-level perspective, which makes it less flexible to re-purpose them for other tasks, but most importantly, demands a lot more data for the model to understand and derive meaningful relationships. We thus operate at the word level rather than sentence-level, to help improve the "sharpness" of the data, with the hope that this would, in turn, help synthesize higher relational features that can offer optimal performance on downstream tasks. To that end, we make some simple assumptions and posit arguments on their choice as better alternatives. 

\subsection{Object vs. Word - a Unit Hypothesis}
The smallest meaningful unit of text is a word, which we assume signifies a visual object embedded in an image. Albeit, the common trend is to train end-to-end network on sentence-level, but system may not be able to learn fine grained contextual relations like \textit{king-queen} analogy. This design choice also aligns with our motivation to generate general-purpose JEs suited for a wide variety of downstream tasks (refer section \ref{sec:eval_imaginator}).

\subsubsection{Number of Objects}

The number of objects in available datasets like Flickr30k \cite{young2014image} and COCO \cite{lin2014microsoft} is limited only to a few hundred. However, if we are interested to learn real-world analogies like \textit{king-queen} analogy, we require far more real-life objects to be detected by the system. Detic \cite{zhou2022detic}, a recent object detection technique, provides 21K object classes and thus, seems the most pertinent. Results shown in  table \ref{tab:allResults} indicate that an increment in the number of objects leads to a corresponding increase the accuracy.

\begin{figure*}[!htp]
\centering
\includegraphics[width=\textwidth]{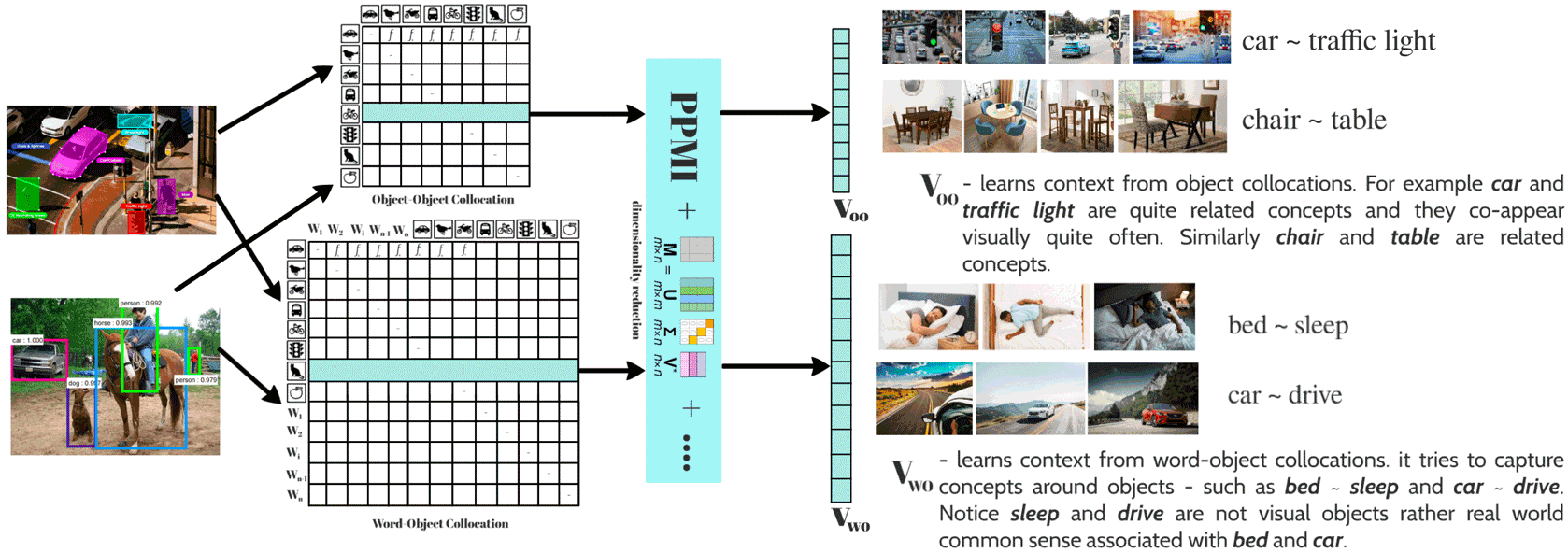}
\includegraphics[width=\linewidth]{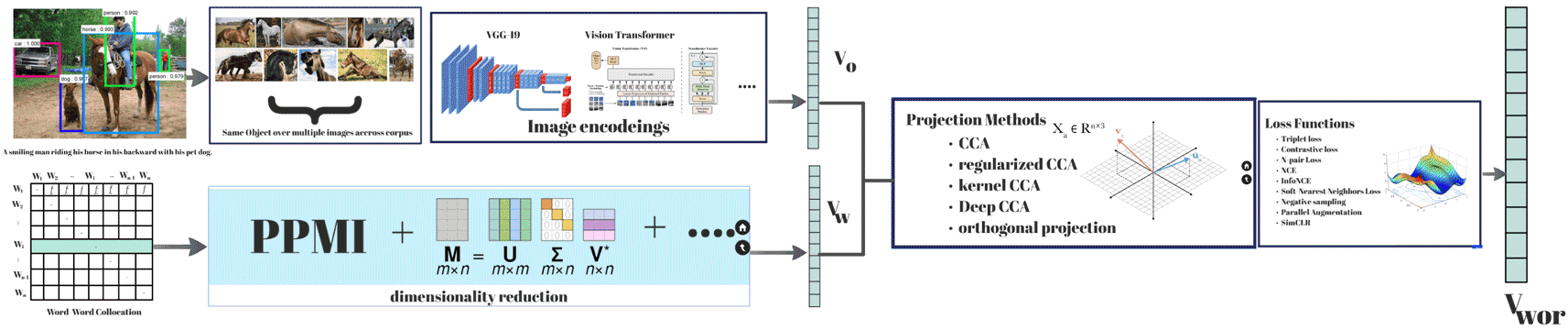}
\caption{ Architecture for creating text embeddings and $v_{oo}$ and $v_{wo}$: the rows and columns in the co-location matrix are the words from the text or objects detected from the images from dataset. Each cell of this matrix represents the occurrence count of each row-column pair in the dataset. The two final vectors are generated using PPMI and eigenvalue weighting over the vectors from co-location matrices. (Bottom) Architecture for learning $v_{wor}$: (left) the averaged VGG19 representation of a particular object across the whole dataset is passed; (right) word2vec representation of the word (i.e., the name of the visual object; for e.g., \textit{horse} in this case).}
\label{fig:textEmbed}
\end{figure*}

Based on the unit hypothesis, we capture three aspects of the input data while generating joint embeddings: Object-object co-location: $v_{oo}$, Word-object co-location: $v_{wo}$, Word-object correlation: $v_{wor}$. 

%% file: 3_lessons_from_NLP.tex
\subsection{Learning $v_{oo}$ and $v_{wo}$}
\label{sec:voo}

Figure \ref{fig:textEmbed} offers a visual summary of the process of generating object-object co-location embeddings $v_{oo}$ and word-object co-location embeddings $v_{wo}$. $v_{oo}$ and $v_{wo}$ are learned using an object co-location matrix, where objects refer to the entities detected using an object detection model. Object co-location matrix is a matrix where the rows and columns correspond to objects detected in our images and each cell represents the co-occurrences of the respective two objects. We then take the rows and apply dimensionality reduction techniques like SVD along with Eigenvalue weighting. The vector obtained is then used as the embedding. This yields \textit{object-object co-location}, which encodes how frequently a detected object co-appears with other detected objects in the dataset. On the other hand, \textit{word-object co-location} is built using the objects from object detection on images and the words from the associated text given in the datasets. This might seem similar to object-object co-location at first glance, but a major difference is that the value in each cell represents the number of image captions having the corresponding object and word pair. With this co-location matrix, we get information on how frequently every object co-appears with other words in the dataset.

%% file: 4_learning_JE.tex
\subsection{Learning $v_{wor}$}
\label{sec:vwor}

Figure \ref{fig:textEmbed} illustrates the process of generating the word-object correlation embeddings $v_{wor}$. 
$v_{wor}$ is learnt using a different approach when compared with the other two embeddings. Co-location can be defined using the co-occurrence of two entities but correlation calls for a deeper understanding of the two entities. Therefore, we get joint embeddings for word-object correlation using word and object vectors.

\begin{figure*}[tp!]
\centering
\begin{minipage}{.42\textwidth}
  \centering
  \includegraphics[width=0.90\textwidth]{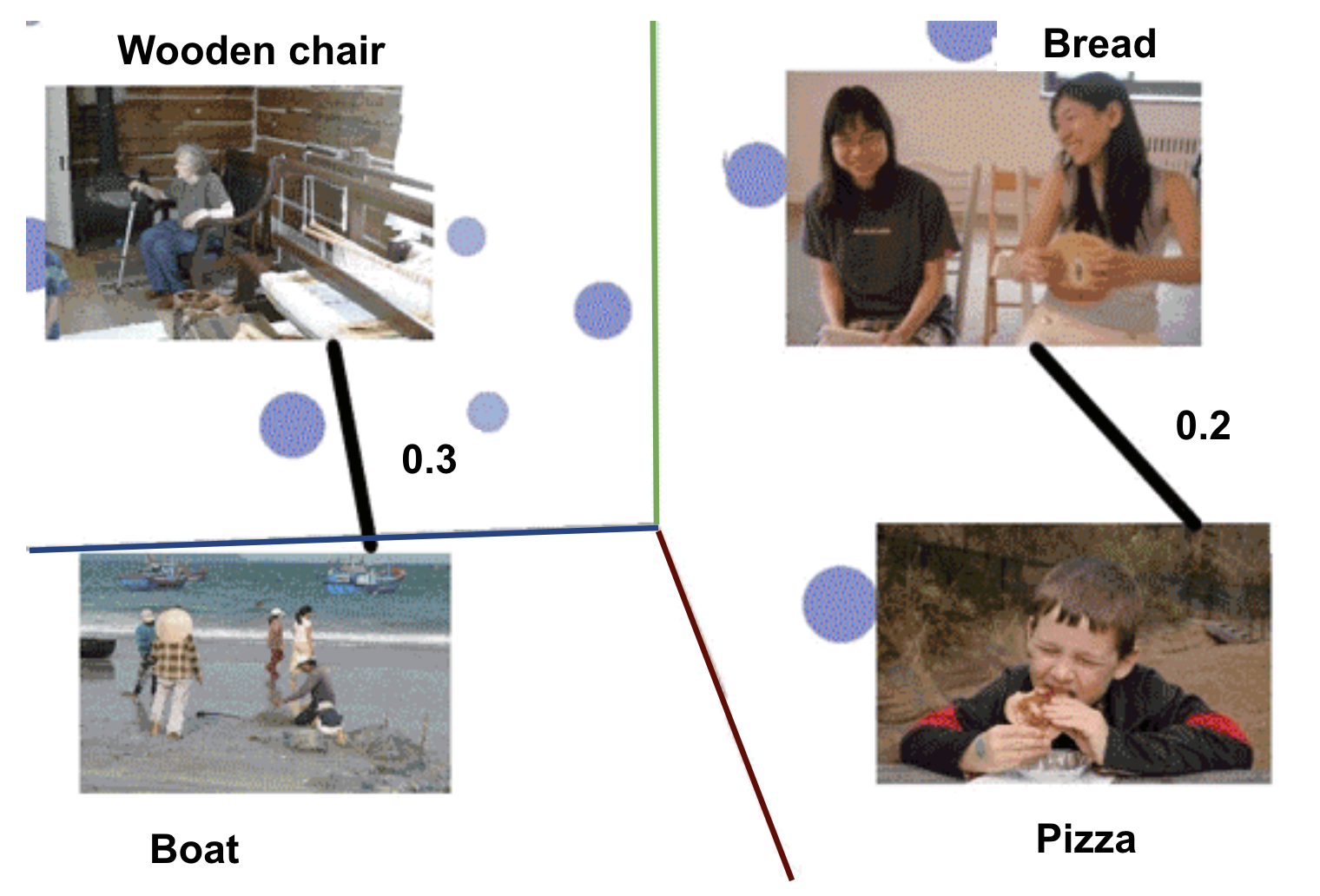}
  \subcaption{}
\end{minipage}
\begin{minipage}{.42\textwidth}
  \centering
  \includegraphics[width=0.90\textwidth]{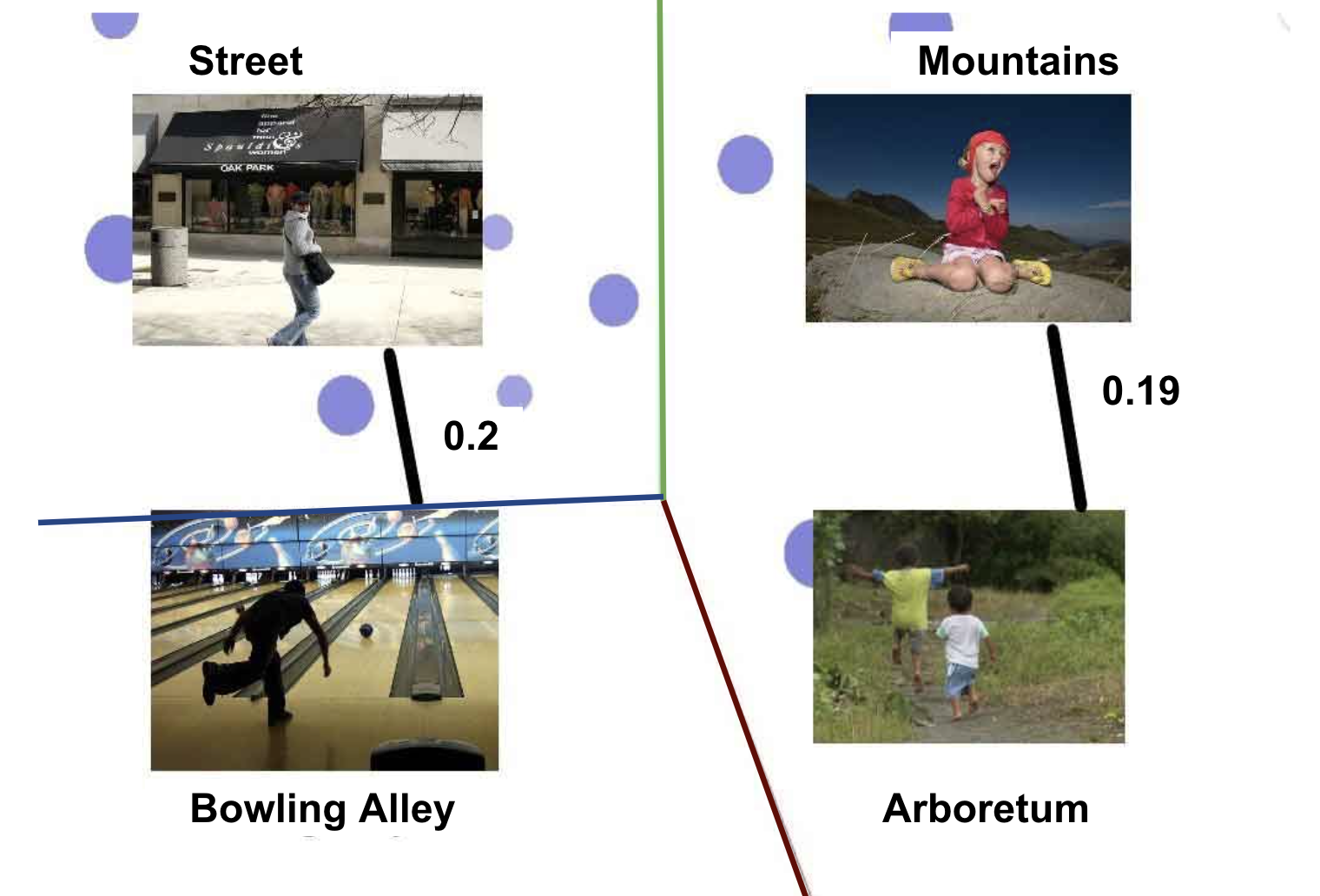}
  \subcaption{}
\end{minipage}
\begin{minipage}{.42\textwidth}
  \centering
  \includegraphics[width=0.90\textwidth]{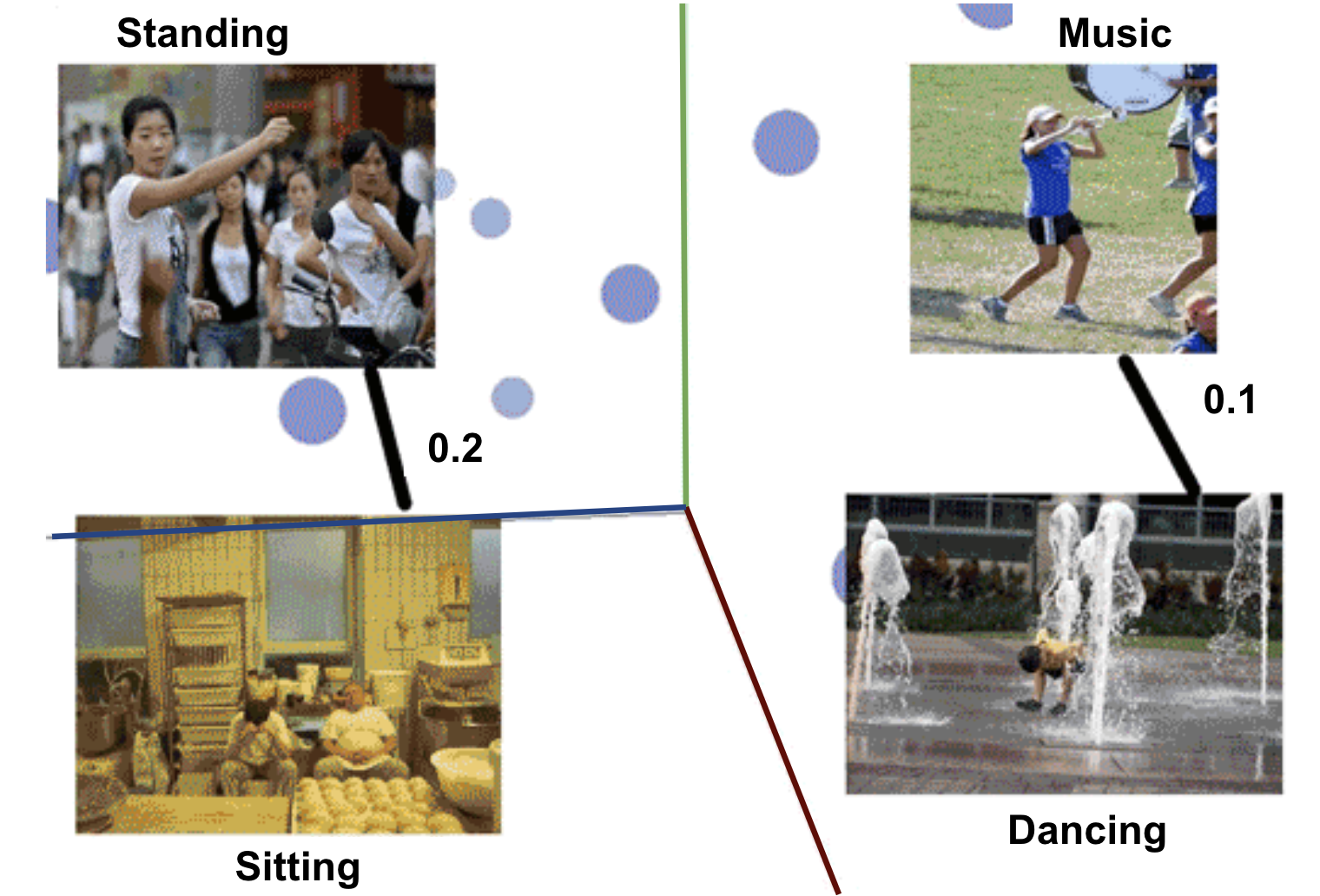}
  \subcaption{}
\end{minipage}
\begin{minipage}{.42\textwidth}
  \centering
  \includegraphics[width=0.90\textwidth]{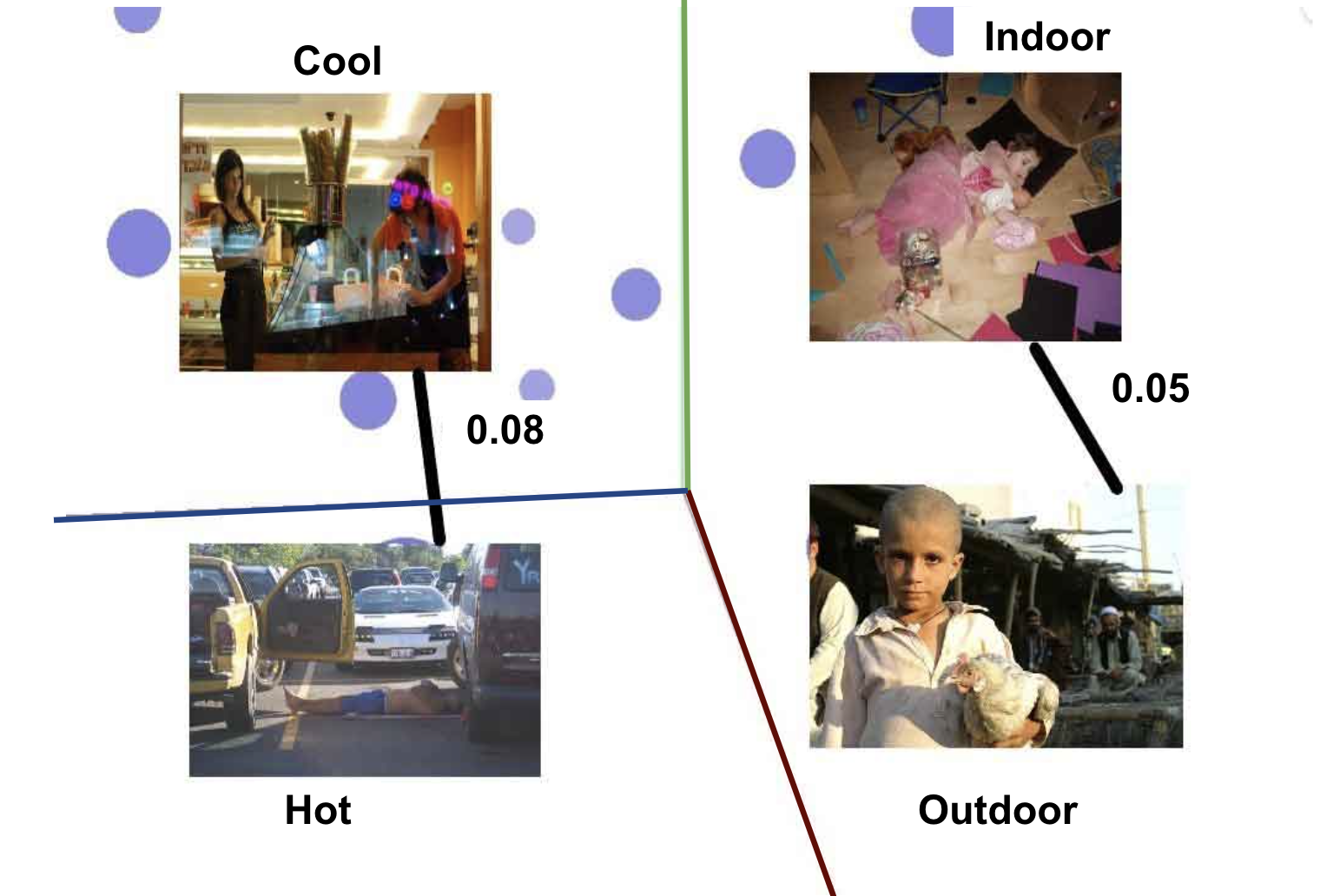}
  \subcaption{}
\end{minipage}
\caption{ Similar images and the vector space distance between them. Word pairs taken from Flickr30k datsset. The IMAGINATOR JE vector space captures real-world analogies well. }
\label{fig:intrinsic}
\end{figure*}

We generate object embeddings by passing all detected crops of the object from the dataset to VGG19 \cite{simonyan2014very}. An average of these embeddings across all instances give us the final embedding for the object encoded as a mean representation. The word embeddings are acquired by creating a \textit{word-word co-location} matrix for the text in the dataset, similar to the aforementioned co-location matrices, where each cell represents the number of co-occurrences of the corresponding word pair.

To obtain the final joint embedding from these two vectors, we project the object embedding in the word embedding space instead of projecting both embeddings in a common space \cite{Kolluru2019ANA, radford2021learning}. The motivation behind this is to maintain the contextuality captured in word embeddings and thus enforce the object embeddings to learn the correlations. Once they learn a correlated vector space, we get the JEs from a weighted average of the projected word and object embedding. 
We perform experiments to compare several projection methods (such as CCA \cite{thompson2000canonical}, Kernel CCA \cite{hardoon2004canonical}, Deep CCA \cite{andrew2013deep} etc.) and loss functions (InfoNCE \cite{oord2018representation}, contrastive loss \cite{hadsell2006dimensionality}, and triplet loss \cite{schroff2015facenet}). Emperically, we find that orthogonal projection and triplet loss give the best JE results. We believe CCA overfits on our data while orthogonal projection \cite{artetxe2018generalizing} uses the features based on the dataset size. Please refer to table \ref{tab:allResults} in Appendix for more on these experiments and their results.

\section{Lessons Learnt from NLP}

\citet{levy2015improving} argue that the performance gains of neural network based word embeddings are due to certain system design choices and hyperparameter optimizations, rather than the embedding algorithms themselves. Furthermore, they show that these modifications can be transferred to traditional distributional models, yielding similar gains. In contrast to prior reports, they show mostly local or insignificant performance differences between the methods, with no global advantage to any single approach over the others. Therefore, we remain grounded to count-based distributional semantics methods. Raw counts or normalized counts are not useful, rather we choose alternatives like PMI and SVD.

\subsection{PPMI and Context Distribution Smoothing}

The PPMI (Positive Pointwise Mutual Information) between a word and its context is well known to be an effective association measure in the word similarity literature. \citet{levy2015improving} show that the skip-gram with negative-sampling training method (SGNS) is implicitly factorizing a word-context matrix whose cell values are essentially shifted PMI. Following their analysis, we present two variations of the PMI (and implicitly PPMI) association metric, which we adopt from SGNS. In this section, $w$ and $c$ represent the word and context matrix.

\textbf{Shifted PMI.} The shift caused by $1<k$ (the number of negative samples in the optimization $(w, c)$: $PMI(w, c) - log(k)$) can be applied to distributional methods through shifted PPMI \cite{levy2014c}:

The $k$ here, firstly, estimates negative sample distribution and secondly, acts as a prior on the probability of an occurrence of $(w,c)$ in the corpus vs. a negative sample. Shifted PPMI captures the latter, i.e, the prior aspect of $k$.

\begin{equation}
    SPPMI(w, c) = max(PMI(w, c) - log(k),\:0)
\end{equation}

\textbf{Context Distribution Smoothing (CDS).} Word2Vec \cite{Mikolov2013EfficientEO} samples negative samples according to a smoothed unigram distribution. This smoothing variation has an analog when calculating PMI directly:

\begin{equation}
    PMI_\alpha(w, c) = log \frac{\hat{P}(w, c)}{\hat{P}(w).\hat{P}_\alpha(c)}
\end{equation}
\begin{equation}
    PMI_\alpha(c) = \frac{\#(c)^\alpha}{\Sigma_c\#(c)^\alpha}
\end{equation}

By enlarging the probability of sampling a rare context (since $\hat{P}_\alpha(c) > \hat{P}(c)$ when $c$ is infrequent), CDS reduces the PMI of $(w, c)$ for a rare context $c$ -- thus removing PMI's bias towards rare words.

\subsection{SVD and Eigenvalue Weighting}
Word and context vectors derived using SVD of co-location matrices can be represented by:
\begin{equation}
    W^{SVD}=U_d\cdot\Sigma_d \;\;\; C^{SVD}=V_d
\end{equation}
However, in this case, $C^{SVD}$ is orthonormal while $W^{SVD}$ is not. Factorization achieved by SGN is much more symmetric and a similar symmetry can be derived using the following factorization:
\begin{equation}
    W=U_d\cdot\sqrt{\Sigma_d} \;\;\; C=V_d\cdot\sqrt{\Sigma_d}
\end{equation}
\citet{levy2015improving} states that while it is not theoretically clear why a symmetric approach performs better for semantic tasks, it works empirically.

For our vector-deriving implementation, we use this as a dimensionality reduction technique. It is similar to SVD but instead of the usual representation: $W=U.\Sigma_d$ and $C=V_d$, eigenvalue weighting uses $W=U.\Sigma_d^{0.5}$ and $C=V_d$. To summarize, after creating the co-location matrix, we derive vectors by initially applying SPPMI with CGS. This is followed by the SVD of the matrices with eigenvalue weighting. 

\subsection{Merging $v_{oo}$, $v_{wo}$, and $v_{wor}$}
The three vectors can be merged using approaches such as concatenation, averaging or autoencoding.  Autoencoder is a pertinent research topic where merging of a number of vectors is learnt automatically by a trained model. This approach considers learning the embeddings by considering complementary information from it's source embeddings.
In the interest of simplifying this aspect of our design, for our experiments, we use weighted average to combine the embeddings. The weights are decided empirically. The best weights we find are 10, 10, and 80 for $v_{oo}$, $v_{wo}$, and $v_{wor}$ respectively. 

%% file: 5_intrinsic.tex
\begin{figure*}[!t]
    \centering
    \includegraphics[width=0.7\textwidth]{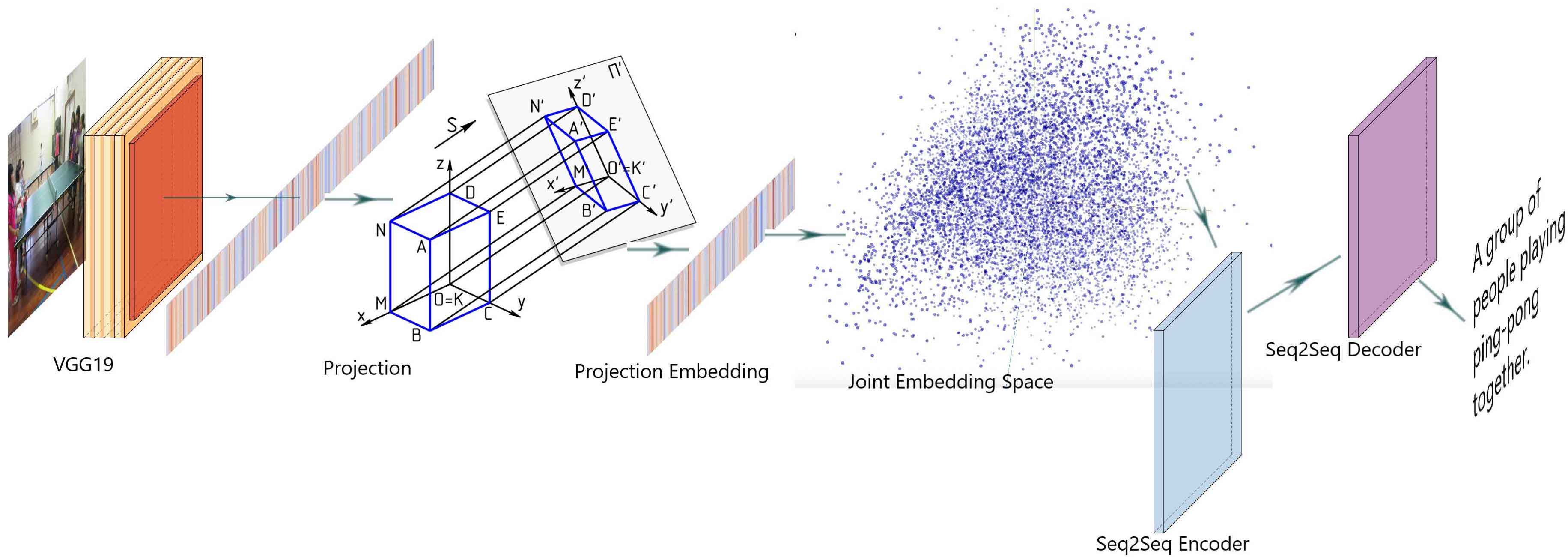}
    \caption{Architecture of Image captioning using IMAGINATOR. We use VGG19 for embedding and then we use IMAGINATOR to project into the joint embedding space. We pick K nearest objects/words and pass to seq2seq model to generate caption.} 
    \label{fig:baselineCaption}
\end{figure*}

\section{Intrinsic Evaluation of IMAGINATOR}
\label{sec:eval_imaginator}
To be able to make vector arithmetic like {\tt King - Man + Woman = Queen} in a generated word vector space is well known as the intrinsic evaluation paradigm. Contemporary image embeddings are devoid of contextuality, whereas text embeddings are much more meaningful, as shown in figure \ref{fig:img1}. With joint embeddings, we aim to add a contextual component to improve the semantic richness of the joint embeddings vector space. We use two kinds of intrinsic evaluation setup to evaluate IMAGINATOR: (i) word contextuality, and (ii) image similarity. 

\begin{table}[htbp!]
\resizebox{0.98\columnwidth}{!}{%
\begin{tabular}{llccc}
\toprule
\textbf{Dataset}                                                 & GloVe & CLiP & SJE  & Ours                           \\ \toprule
WS353 \cite{finkelstein-2001-placing}           & 2.65  & 0.35 & 0.19 & 0.14                           \\
MTurk \cite{10.1145/2339530.2339751}            & 1.99  & 0.26 & 0.28 & 0.20                           \\ 
RG65 \cite{10.1145/365628.365657}               & 0.75  & 0.38 & 0.20 & 0.13                           \\ 
RW \cite{pilehvar-etal-2018-card}               & 0.96  & 0.19 & 0.17 & 0.25                           \\ 
SimLex999 \cite{hill-etal-2015-simlex}          & 2.31  & 0.18 & 0.22 & 0.12                           \\ 
MEN \cite{10.5555/2655713.2655714}              & 0.51  & 0.22 & 0.13 & 0.11                           \\ 
Google Analogy \cite{Mikolov2013EfficientEO}    & 2.09  & 0.18 & 0.12 & 0.15                           \\ 
MSR Analogy \cite{mikolov-etal-2013-linguistic} & 0.63  & 0.30 & 0.09 & 0.22                           \\
SemEval2012 \cite{jurgens-etal-2012-semeval}    & 1.2   & 0.21 & 0.32 & 0.26                           \\ 
BLESS \cite{baroni-lenci-2011-blessed}          & 2.77  & 0.22 & 0.19 & 0.11                           \\ \bottomrule
Average                                                          & 1.5   & 0.25 & 0.19 & \textbf{0.16} \\ 
\bottomrule       
\end{tabular}
}
\caption{Results (average euclidean distance) for intrinsic evaluation of our JEs based on notable word contextually  datasets. Lower is better. Our model outperforms the baselines on most of the datasets and has the lowest overall average distance. }
\label{tab:WordAnalogy}
\end{table}

\subsection{Word Contextuality}

We  use all the $10$ datasets mentioned in \cite{DBLP:journals/corr/JastrzebskiLC17} to evaluate the generated word embeddings intrinsically. Intrinsic means that only basic arithmetic functions are performed on the embeddings and no other models are trained. The datasets cover three tasks: (i) word similarity, (ii) word analogy, and (iii) word categorization. First, the word embeddings for given pair of similar words from the datasets are computed. Then, we use average euclidean distance to derive the final results ( as shown in table \ref{tab:WordAnalogy}) for embeddings from GloVe \cite{pennington-etal-2014-glove}, CLIP \cite{radford2021learning}, and IMAGINATOR. We can see that IMAGINATOR performs better than the baselines.

\subsection{Image Similarity}

Analogy-making on images is relatively challenging. Our hypothesis is that vectors of the same/similar objects must be nearby in the IMAGINATOR vector space. We evaluate IMAGINATOR intrinsically on image similarity task using objects five datasets - Caltech 101 \cite{caltech101}, Flickr 30k, MS COCO, Google CC \cite{sharma-etal-2018-conceptual}, Visual Genome \cite{krishnavisualgenome}.  We extract the list of similar objects from the the datasets, obtain features from the VGG19 and then orthogonally \cite{artetxe2018generalizing} project those objects to the IMAGINATOR vector space. We then calculate the pairwise-euclidean distance between such vectors and average them for the entire dataset. Table \ref{tab:ImageSimilarity1} shows the object similarity performance of SJE, CLIP and IMAGINATOR on a a variety of datasets. Our baseline comprehensively outperfroms the baselines on all the datasets.  Figure \ref{fig:intrinsic} shows some examples of the relation between projected JEs of these objects. From the examples we can see that IMAGINATOR captures the nuances of the images.

\begin{table}[htbp!]
\centering
\resizebox{0.8\columnwidth}{!}{
\begin{tabular}{lccc}
\toprule
Dataset       & SJE & CLIP  & IMAGINATOR \\* \toprule
Caltech 101   &  1.9 &  1.5    & 0.13  \\
Flickr 30K    &   0.8  &  0.4   &  0.06          \\
MS COCO       &  0.9 & 1.3    & 0.2            \\
Google CC     &   0.2  &  0.4&  0.08          \\
Visual Genome &    1.1 &  1.4 & 0.1          \\*
\midrule
Average      &  0.98 & 1.00 &  \textbf{0.11}
\\
\bottomrule
\end{tabular}%
}
\caption{Average pairwise euclidean distance between similar objects from each dataset. Lower is better.}
\label{tab:ImageSimilarity1}%
\end{table}

%% file: 6_extrinsic.tex
\section{IMAGINATOR for Downstream Tasks}
\label{sec:image_captioning}

\begin{figure*}[htbp!]
\centering
\begin{subfigure}{0.5\textwidth}
  \centering
   \includegraphics[width=0.45\textwidth]{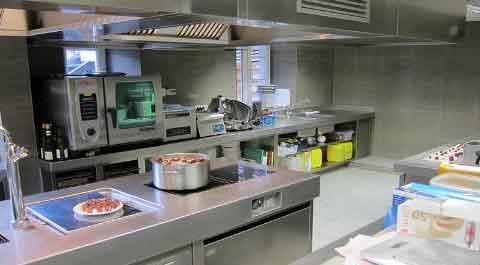}
    \caption{\centering       \textbf{IMAGINATOR}: A kitchen with a sink, stove, oven, and beers.
    \textbf{Gold Caption}: A commercial stainless kitchen with a pot of food cooking.
   \textbf{OSCAR}: a kitchen with a lot of pots and pans in it.
}
\end{subfigure}%
\begin{subfigure}{0.5\textwidth}
  \centering
   \includegraphics[width=0.4\textwidth]{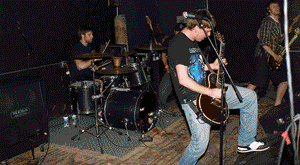}
    \caption{ \centering  \textbf{IMAGINATOR}: A vocalist, drummer, and a guitarist sings a tune. \hspace{\textwidth}
    \textbf{Gold Caption}: A musical band are by their instruments most likely playing a song. \hspace{\textwidth} \textbf{OSCAR}: A man on a stage with a guitar and a keyboard.
    }
\end{subfigure}%

\caption{Examples of some image captioning outputs generated by IMAGINATOR along with the original caption and the caption generated by OSCAR \cite{li2020oscar}. IMAGINATOR gives richer and more detailed captions than OSCAR. For more examples please refer the appendix.}
\label{fig:captioningExamples}
\end{figure*}

The downstream vision-language (VL) tasks chosen to test our pre-trained JEs are: (i) image captioning, (ii) Image2Tweet \cite{jha-etal-2021-image2tweet}, and (iii) text-based image retrieval.

\subsection{Image Captioning}
Image captioning is a common multimodal task which involves the generation of a textual description for an image. Describing the contents of an image requires visual understanding at an object level. We use JEs from IMAGINATOR to generate captions on datasets such as Flickr30k \cite{young2014image} and COCO \cite{lin2014microsoft}.

\begin{figure*}[htbp!]
\centering
\begin{subfigure}{0.35\textwidth}
  \centering
   \includegraphics[width=0.55\textwidth]{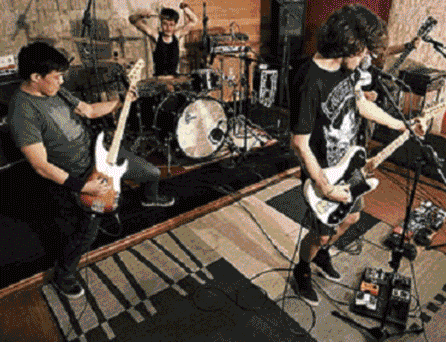}
    \caption{ rank 1 image
    ALBEF \cite{albef}
    }
\end{subfigure}%
\begin{subfigure}{0.33\textwidth}
  \centering
   \includegraphics[width=0.48\textwidth]{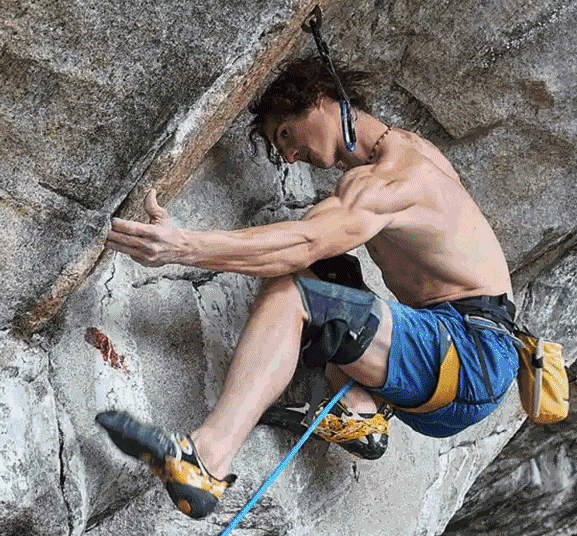}
    \caption{ rank 1 image 
    XVLM \cite{xvlm}
    }
\end{subfigure}%
\begin{subfigure}{0.34\textwidth}
  \centering
   \includegraphics[width=0.45\textwidth]{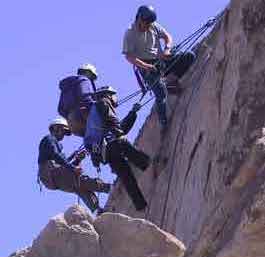}
    \caption{ rank 1 image
    BERT$_{IMAGINATOR}$
    }
\end{subfigure}%

\caption{Image retrieved for the query - \textit{"several climbers climbing rock together"} - it is evident that ALBEF \cite{albef} wrongly emphasized on \textit{"rock together"}, whereas XVLM \cite{xvlm} is unable to comprehend plurality in the query here, while BERT$_{IMAGINATOR}$ can do the job well.}
\label{fig:IR_Captions_main}
\end{figure*}

For an input image, we start by obtaining an image embedding using VGG19 \cite{simonyan2014very}, which is then orthogonally projected in IMAGINATOR embedding space. We use the JE of the image to find $k$ nearest objects in the vector space. For our experiments we used $k=10$, giving us 10 objects associated with the input image. These objects are then passed to a sequence-to-sequence module, namely, the T5 transformer \cite{bhatia}, which generates the final caption. We use a pre-trained T5 model, fine-tuned on Flickr30k and COCO. Figure \ref{fig:baselineCaption} describes the captioning pipeline while Figure \ref{fig:captioningExamples} shows some output examples. 

Table \ref{table:tab1} shows the quantitative results of baseline models and IMAGINATOR. We can see that our model outperforms all the baselines in terms of BLEU score and BERTScore \cite{zhang2020bertscore} on both the datasets.

\begin{table}[htbp!]
\centering
\resizebox{0.98\columnwidth}{!}{%
\begin{tabular}{lllll} \toprule
method                              & \multicolumn{2}{l}{Flickr30k} & \multicolumn{2}{l}{COCO}      \\ \toprule
                                    & BLEU          & BERTScore     & BLEU          & BERTScore     \\ \toprule
UVLP                                & 30.1          & -             & -             & -             \\
OSCAR                               & -             & -             & 41.7          & -             \\
SJE         & 30.5          & 0.78          & 35.6          & 0.8           \\
CLIP        & 31.3          & 0.83          & 36.3          & 0.85          \\
BLIP                                & -             & -             & 40.4          & -             \\
IMAGINATOR & \textbf{33.2} & \textbf{0.87} & \textbf{43.1} & \textbf{0.88} \\ \bottomrule
\end{tabular}%
}
\caption{Results of different models on the image captioning task. Higher is better. Unavailable scores are left blank.  }
\label{table:tab1}
\end{table}

\subsection{Image2Tweet}
\label{sec:image2tweet}
Image2Tweet \cite{jha-etal-2021-image2tweet} is a task which is a notch above traditional image captioning in terms of complexity. Given an input image, the task involves generating a tweet like a human news reporter. Figure \ref{fig:moreimg2tweet_examples} shows some examples from the dataset.

The tweet is generated using a method similar to image captioning. The joint embedding of the input image is used to find the $k$ nearest neighbouring embeddings in the projections space. These neighbours are then used to generate the tweet using a sequence-to-sequence model.

\begin{figure*}[htbp!]
    \centering
     \includegraphics[width=0.65\textwidth]{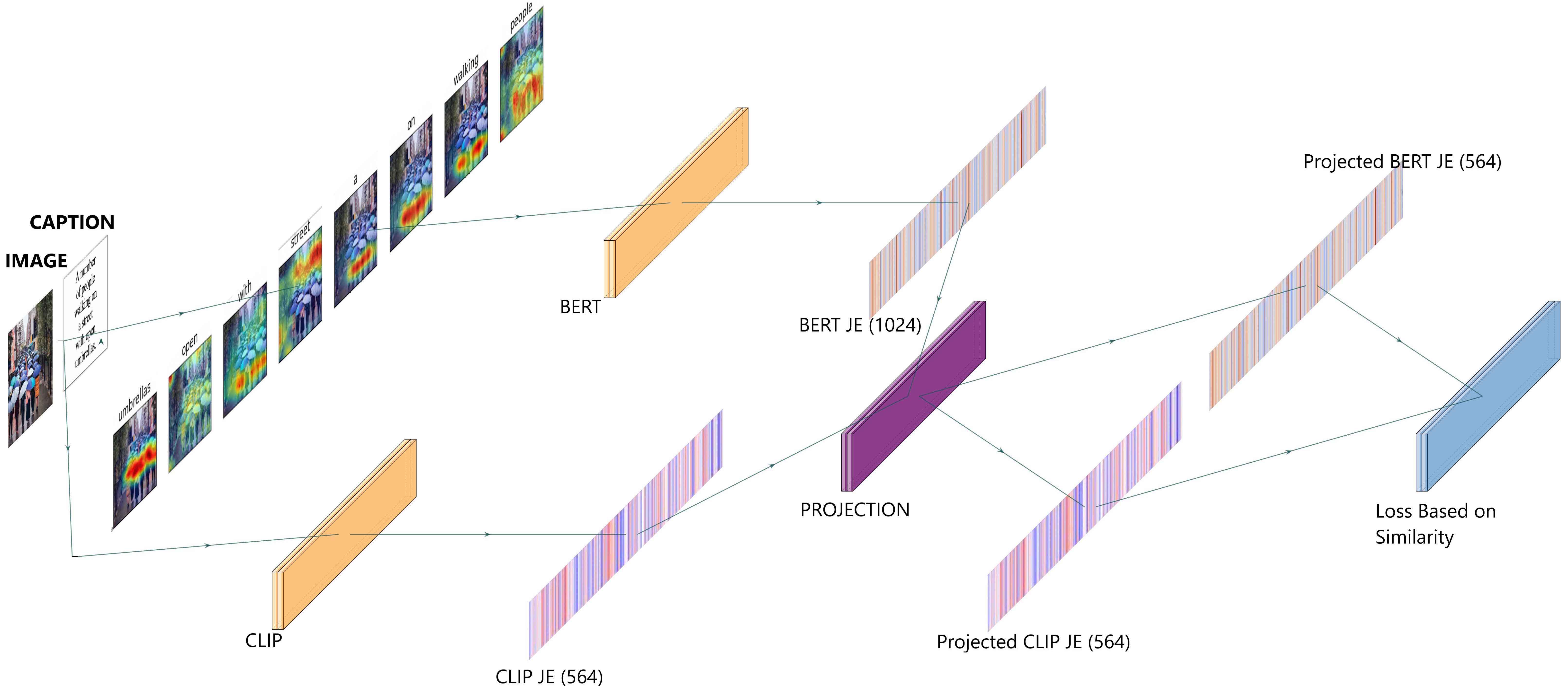}
    \caption{$BERT_{IMAGINATOR}$ - Training approach for Image Retrieval. Training happens in batches and cosine similarity between corresponding image-sentence pair is maximised while for other pairs it is minimized.}
    \label{fig:bertje_training}
\end{figure*}

The results are based on the CIDEr metric (refer table \ref{table:img2tweet}). We found that using other datasets for training SoTA models yielded abysmal results, indicating that Image2Tweet is a fairly complex problem. However, IMAGINATOR performs reasonably well on the task and surpasses comparable contemporary SoTA captioning methods, namely UVLP \cite{zhou2020unified} and OSCAR \cite{li2020oscar}.

\begin{table}[!hb]
\centering
\resizebox{0.9\columnwidth}{!}{%
\begin{tabular}{lc} 
\toprule
Method & CIDEr \\
\toprule
Baseline of Image2tweet \cite{jha-etal-2021-image2tweet} & 0.0003 \\
UVLP \cite{zhou2020unified} [SoTA on Flickr] & 0.003 \\
OSCAR \cite{li2020oscar}  [SoTA on COCO] & 0.004\\
CLIP \cite{radford2021learning} & 0.006\\
Stanford joint embedding \cite{Kolluru2019ANA} & 0.007 \\
5 ensemble \cite{luo2018discriminability} & 0.0090 \\
IMAGINATOR + $k$ nearest objects + T5 & \textbf{0.0095} \\
\bottomrule
\end{tabular}
}
\caption{Results (CIDEr score) of various multi-modal models on the Image2Tweet task. Higher is better. Our model outperforms other all models.
}
\label{table:img2tweet}
\end{table}

\subsection{Text-based Image Retrieval}
\label{sec:image-retrieval}

The fundamental question that we are seeking an answer to is whether using IMAGINATOR word-object level embeddings we can achieve compositionally and achieve a vector representation for sentence-image level. For example, by passing on word vectors in a sequence to a language model we can obtain a sentence-level vector representation. To verify the compositionality of joint modality embeddings, we test our approach on the task of text-based image retrieval on the Flickr30K dataset \cite{young2014image}. The main challenge of this task is to find out the appropriate content in the visual space while the input is in the text space. Another reason for introducing compositionality is that each word is usually associated with multiple images. Hence, there is a need for us to learn a single image representation for a given text. Though we explore contrastive methods in section \ref{contrastivelearning}, to solve the above-mentioned challenges, we introduce an approach using BERT and evaluate it on text-based image retrieval.

\begin{table}[!ht]
 \centering
 \resizebox{0.8\columnwidth}{!}{%
 \begin{tabular}{lccc} 
 \toprule
 Method & R@1  & R@5 & R@10  \\
 \toprule
 ALBEF \cite{albef} & 85.6 & 97.5 & 98.9   \\
 XVLM \cite{xvlm} & 86.1  & 97.3 & 98.7 \\
 BLIP \cite{li2022blip} & 87.6 & 97.7 &99.0\\
 BERT$_{IMAGINATOR}$ & \textbf{89.48} &\textbf{98.1} &\textbf{99.2} \\
 \bottomrule
 \end{tabular}
 }
 \caption{Results on image retrieval: Recall@\{1, 5, 10\} Score for the Flickr30K dataset. Higher is better.}
 \label{tab:IR}
 \end{table}

\subsubsection{Compositionality of Joint Embeddings -  \texorpdfstring{$BERT_{IMAGINATOR}$}{Lg}}

BERT is arguably the most successful modelling architectures in NLP. It accepts token embeddings as input and produces contextualized embeddings as output. In contrast, we propose $BERT_{IMAGINATOR}$, which is trained to take image+text as input and output a compositional vector representation for both modalities.

We utilize BERT \cite{devlin2018bert} and CLIP \cite{radford2021learning} as our backbones to generate JEs. 
Instead of feeding the BERT model tokenized words obtained via a tokenizer, we use IMAGINATOR (refer section \ref{sec:vwor}) word-object embeddings as input to the model. We process necessary tokenization, position encoding, and segment embeddings accordingly, per the BERT architecture.

We utilize CLIP \cite{radford2021learning} for generating another JE using an image-sentence pair by obtaining the image and text embeddings from CLIP encoders and concatenating them. We refer to this as the \textit{sentence JE}. Both these embeddings, viz., the \textit{sentence JE} and \textit{projected $BERT_{IMAGINATOR}$}, are projected to a common space using orthogonal projection \cite{artetxe2018generalizing}, on which we compute our loss. Figure \ref{fig:bertje_training} visually depicts our training process while table \ref{tab:IR} shows BERT$_{IMAGINATOR}$ outperforming SoTA information retrieval (IR) baselines, namely ALBEF \cite{albef} and XLVM \cite{xvlm} on Recall@\{1, 5, 10\}. Some output examples are shown in figure \ref{fig:IR_Captions_main}.

%% file: 8_conclusion.tex
\section{Conclusion and Futurework}

We proposed a new pre-trained joint embedding IMAGINATOR. Our major contribution is on adopting count-based methods for joint modality, echoing the philosophy from \citet{levy2015improving}. We present an in-depth intrinsic evaluation along with a new architecture $BERT_{IMAGINATOR}$. IMAGINATOR outperformed SoTA on three tasks: \textit{(i) image captioning, (ii) Image2Tweet, and (iii) text-based image retrieval}. In the future, we would like to explore other multimodal tasks such as VQA.

\section*{Discussion and Limitations}

While IMAGINATOR pushes the boundaries of the state-of-the-art in tasks that involve language and vision joint modelling, there are some limitations.

\subsection*{Object Detection - Limited Number of Classes}
IMAGINATOR utilizes the atomic units of multimodal data -- individual words for text representation and individual objects for image representation. Typically, the number of unique words (i.e., the vocabulary) is quite large in a given text relative to the number of objects in images. As such, IMAGINATOR being a joint learning technique is bottlenecked by the capabilities of existing object detection techniques since they only typically deal with a limited repertoire of objects. 
To enhance the richness and expressivity of JEs, object detection models that can identify the wide gamut of objects in the world would be critical.

\subsection*{Contrastive Learning}
\label{contrastivelearning}
Contrastive learning is a task-independent technique that focuses in learning the similarity and differences between samples in a dataset. The objective here is to learn an embedding space where similar inputs, say samples belonging to the same class, are embedded as similar representations while samples from dissimilar classes are separated in the embedding space. IMAGINATOR performs well in several tasks, despite out our object representation being a simple average of image embeddings. However, contrastive learning might be able to learn even better vectors that capture the relations between images and their objects.

\subsection*{Vision Transformer and Positional Encoding}

A Vision Transformer (ViT) is a transformer that is targeted at vision processing tasks, such as object recognition and is much more robust than CNNs. It divides an image into fixed-size patches, embeds each of them, and includes a positional embedding along with the patch embedding as an input to the transformer encoder. In our case, if we could draw meaningful cross-modal connections between sections of text and the corresponding parts of images, a significant performance uptick can be potentially reached. This can be implemented using the various positional encoding schemes in ViT.

%% file: 9_appendix.tex
\newpage
\section*{Appendix} 
\label{appendix}

\subsection*{What is the value-addition of this work given Joint Embeddings have been explored in various ways for the past few years?}
\label{appendix:related}
\begin{figure*}[htp]
    \centering
    \includegraphics[width=0.75\textwidth]{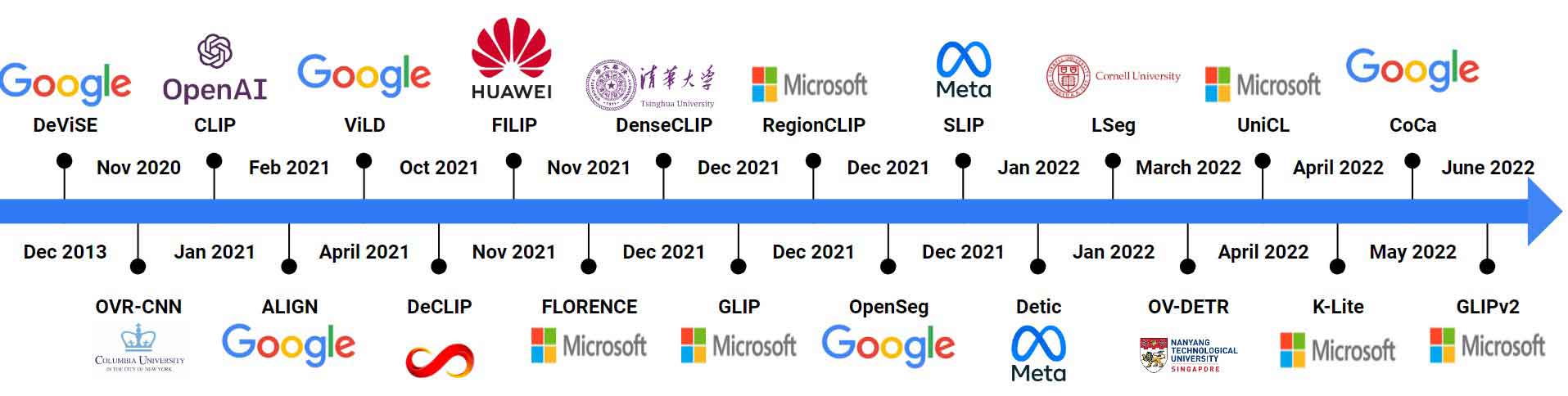}
    \caption{Notable recent work related to vision-language pre-training. Taken from \citet{cvprtut}.}
    \label{fig:img2}
\end{figure*}

Learning joint embeddings has been a topic that has received immense interest from the multimodal AI community over the past decade \cite{chen2020uniter, jia2021scaling, tan2019lxmert}. A concise survey on this topic has been presented in \citet{cvprtut}, which offers an extensive treatment of both early (i.e., input-level), mid (i.e., feature-level), and late (i.e., decision-level) fusion methods, depicted visually in figure~\ref{fig:img2}. Learning joint embeddings using early-fusion methods (like the one we adopted in our work) essentially enables identifying cross-correlations between various modalities (such as text, images, video, audio, spatial/point-cloud information, etc.) early on in the learning process. As such, the resultant vector representations typically lead to top-notch performance in most downstream tasks. On the other hand, a vast majority of work focuses on feature fusion where modalities are first individually processed and then projected to a common vector space to draw correlations using variety of projection methods like CCA \cite{thompson2000canonical}, Deep CCA \cite{andrew2013deep}, etc.

As mentioned in Section \ref{sec:related_work}, IMAGINATOR's novelty is associated with the word-level grounding of objects using traditional count-based approaches, an NLP tradition that was prevalent before the neural era. This is a significant detour from recent work in learning joint embeddings that uses deep learning-based techniques, which suffer from a lack of control or ease of interpretation owing to their inherent black-box nature. As such, this design decision has allowed us to learn rich features that are co-location-based representations of visual objects that are grounded in words which represent the object's moniker(s). The co-location-based contextual word vectorization is primarily influenced by the distributional hypothesis \textit{"You shall know a word by the company it keeps"} \cite{firth1957}. Intrigued by how such a co-location-based method can aid visual contextual learning, we sought to testify its utility in learning joint emebddings. However, the ability of count-based joint embedding techniques can be severely limited due to the insufficient number of objects detected, which led to us overcoming this issue by using statistical correlational methods inspired by NLP (\citet{levy2015improving}. We plan to further scale this technique by first enabling detection of additional (>20K) visual objects, hypothesizing that this learning paradigm can lead to even richer representations.

\clearpage
\subsection*{Rolodex of additional experiments carried out for the optimal generation of $v_{oo}$, $v_{wo}$, and $v_{wor}$} 

Table \ref{tab:allResults} expands on section ~\ref{sec:voo} and ~\ref{sec:vwor} and shows a comparison of different embedding methods, dataset combinations, number of objects, loss functions, and projection methods with performance on the captioning task. We consider normalized count, PMI, PPMI, and Factorized PPMI for vector building and SVD and Eigenvalue factorization for dimensionality reduction. 
For projection, we consider Orthogonal Projection, CCA, regularized CCA, and Deep CCA for our experiments. From Table \ref{tab:allResults}, we can see that with an increase in the number of objects, the captioning score also correspondingly increases.

\begin{table*}[!htbp]
\centering
\begin{adjustbox}{angle=0}
\resizebox{\textwidth}{!}{
    \begin{tabular}{lccccccc}
    \toprule
    \multirow{2}{*}{Embedding method} & \multirow{2}{*}{Dataset} & \multirow{2}{*}{No. of objects} & \multirow{2}{*}{Loss function} & \multirow{2}{*}{Projection method} & \multicolumn{2}{c}{Performance} \\
    & & & & & Flickr30K & COCO\\ \toprule 
    Normalized count     & Flickr30K & 1000  & Triplet loss & Orthogonal      & 32.1 & 29.3 \\
                     & Flickr30K + COCO             & 1080  & Triplet loss & Orthogonal      & 32.4 & 33.7 \\
PMI                  & Flickr30K + COCO     & 17000 & Triplet Loss & -               & 33.4 & 34.8 \\ \hline
PPMI                 & Flickr30K + COCO     & 17000 & Triplet Loss & -               & 33.9 & 35.4  \\ \hline
Factorized PPMI      & Flickr30K + COCO     & 17000 & Triplet Loss & -               & 34.1 & 37.9 \\ \hline
Factorized PPMI + SVD & Flickr30K + COCO     & 17000 & Triplet Loss & -               & 32.4 & 36.2 \\ \hline
\multirow{3}{2cm}{Factorized PPMI + Eigen Value Factorization} & \multirow{3}{*}{Flickr30K} & \multirow{3}{*}{1000} & \multirow{3}{*}{Triplet Loss} & CCA & 30.2 & 32.2 \\
 & & & & Regularized CCA & 30.9 & 33.9 \\
 & & & & Deep CCA & 30.5 & 33.2 \\
                     & Flickr30K + COCO & 1080  & Triplet loss & Orthogonal & 31.9 & 35.7  \\
                     & Flickr30K + COCO     & 17000 & Triplet loss & Regularized CCA & 33.2 & 40.1  \\
                     & Flickr30K + COCO     & 17000 & Triplet loss & Deep CCA        & 33.8 & 38.1   \\ 
                     \bottomrule
    \end{tabular}
}
\end{adjustbox}
\caption{Results on image captioning datasets (Flickr30K and COCO) for different embedding methods, datasets, loss functions, and projection methods.
}

\label{tab:allResults}
\end{table*}

\newpage

\clearpage

\subsection*{Intrinsic evaluation of IMAGINATOR}\label{sec:app-int-eval}
\begin{figure*}[htp!]
\centering
\begin{minipage}{.45\textwidth}
  \centering
  \includegraphics[width=\textwidth]{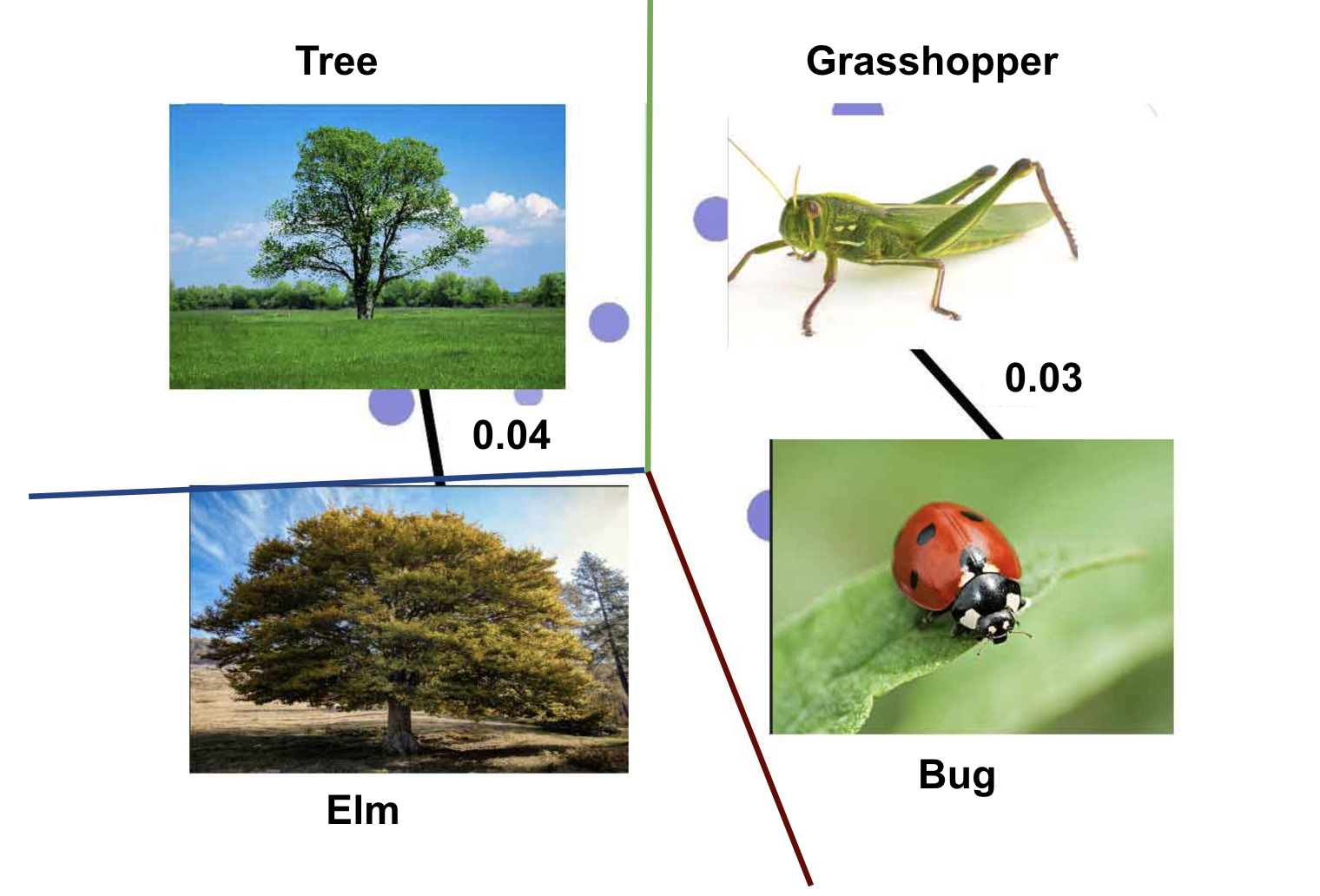}
  \subcaption{}
\end{minipage}
\begin{minipage}{.45\textwidth}\raggedleft
  \centering
  \includegraphics[width=\textwidth]{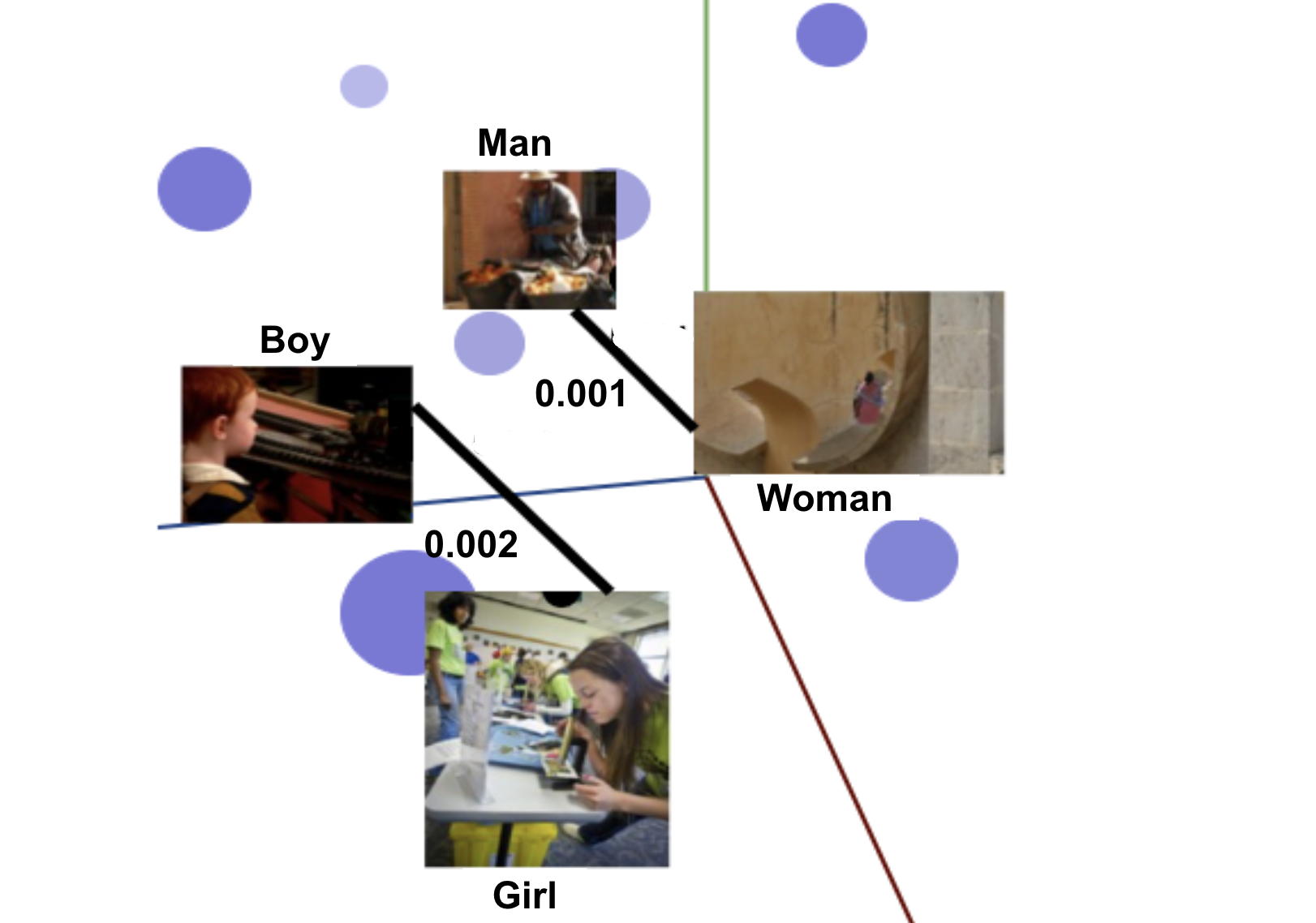}
  \subcaption{}
\end{minipage}

\begin{minipage}{.45\textwidth}\raggedleft
  \centering
  \includegraphics[width=\textwidth]{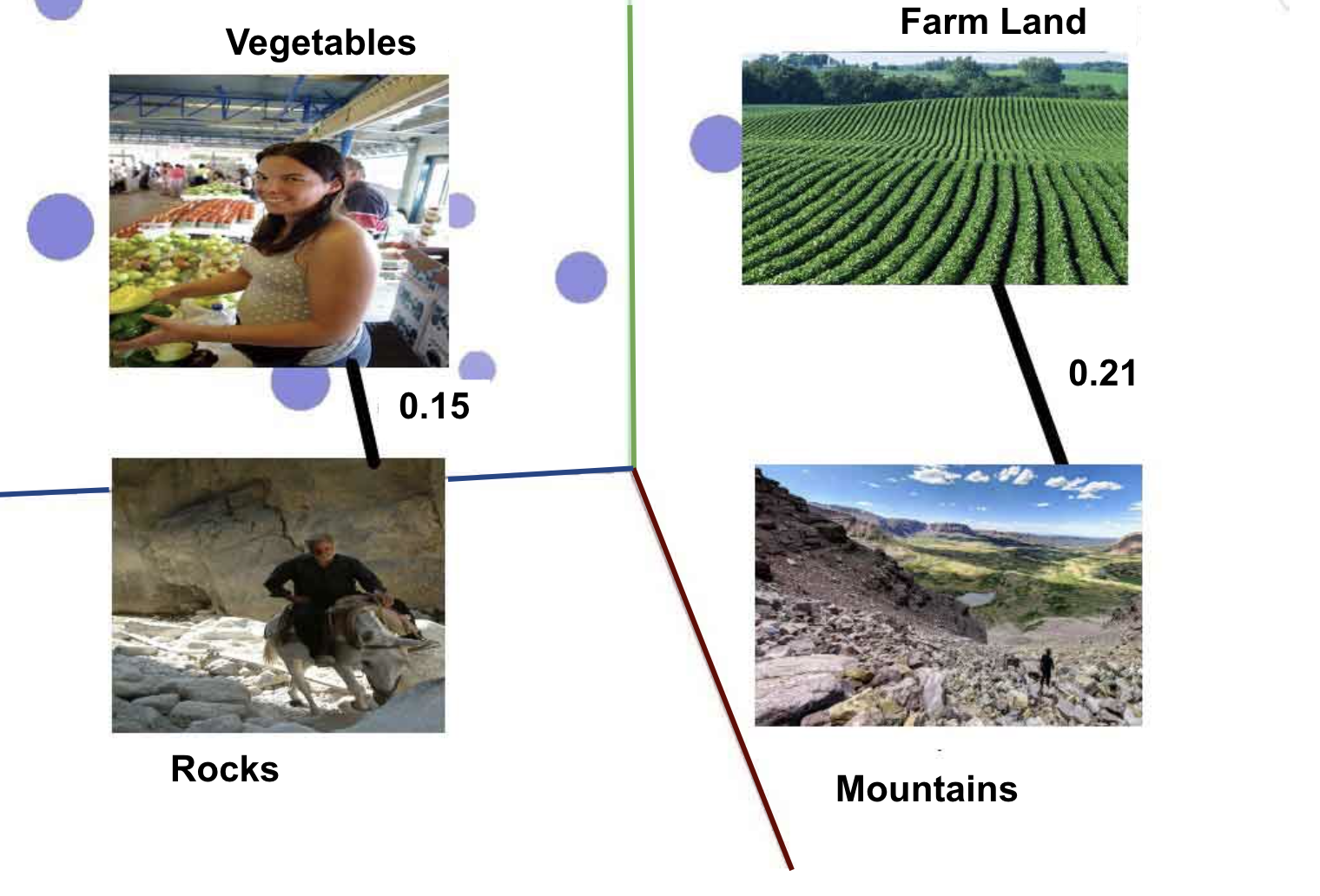}
  \subcaption{}
\end{minipage}
\begin{minipage}{.45\textwidth}\raggedleft
  \centering
  \includegraphics[width=\textwidth]{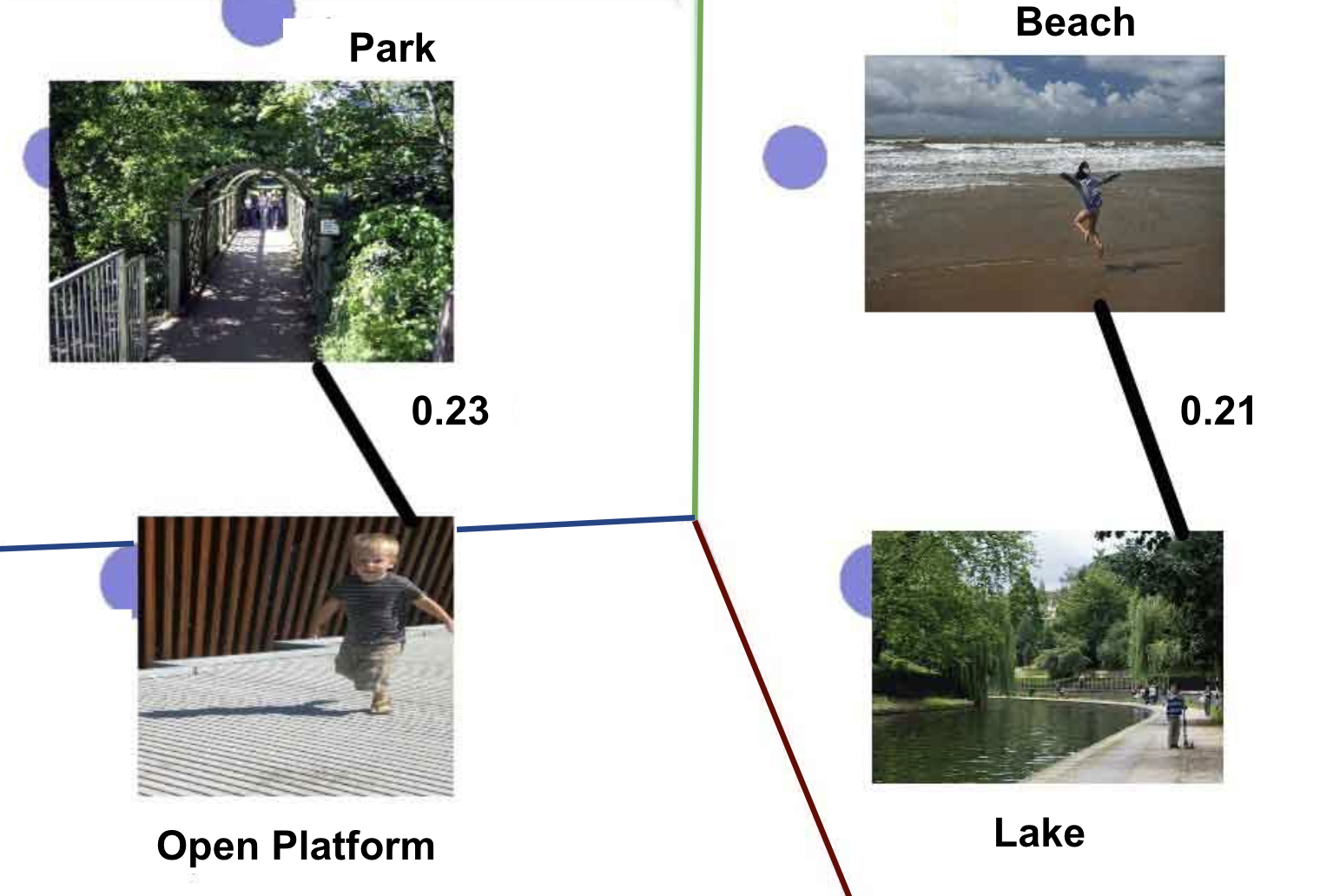}
  \subcaption{}
\end{minipage}

\begin{minipage}{.45\textwidth}\raggedleft
  \centering
  \includegraphics[width=\textwidth]{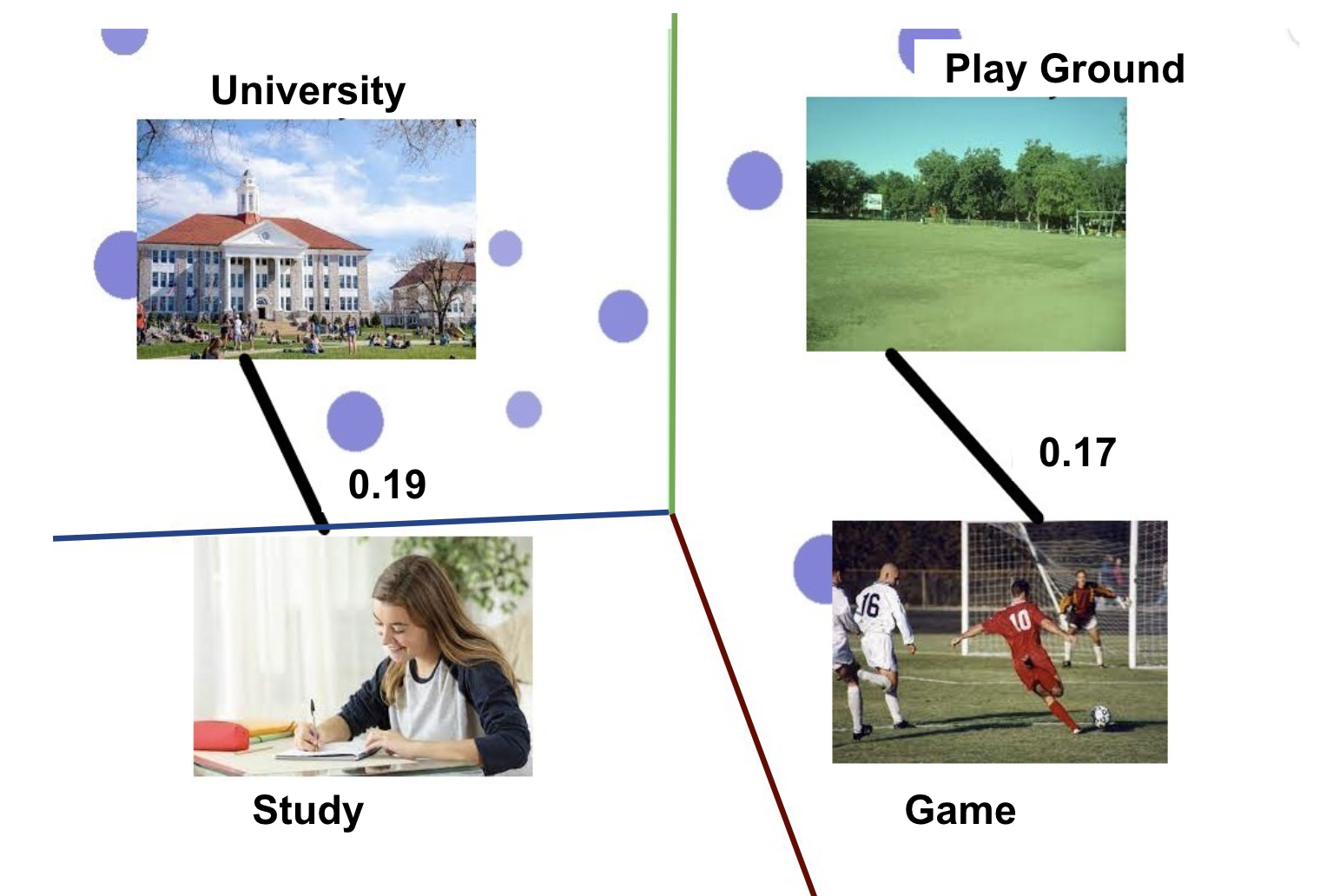}
  \subcaption{}
\end{minipage}
\begin{minipage}{.45\textwidth}\raggedleft
  \centering
  \includegraphics[width=\textwidth]{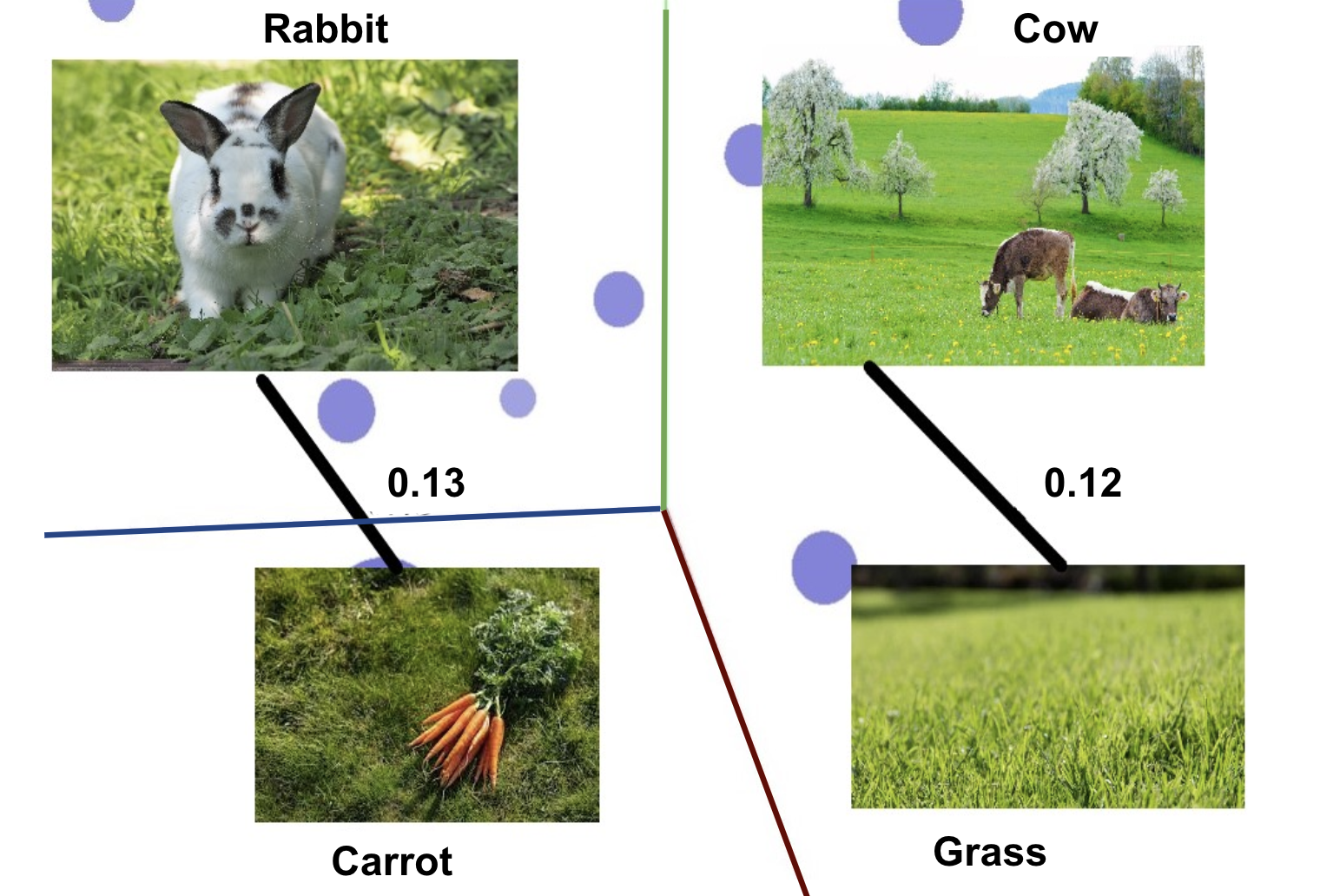}
  \subcaption{}
\end{minipage}

\caption{Examples for intrinsic evaluation of our JEs based on  image similarity.}
\label{fig:moreInstrinsic}
\end{figure*}

The goal behind intrinsic evaluation is to understand how well the embeddings adhere to the contextuality constraint. Building upon section ~\ref{sec:eval_imaginator}, we consider standard relational terms - king, queen, boy, woman and performed an intrinsic evaluation on them to identify the relationships between these terms. We project the joint embeddings of the image and check the Euclidean distance among them; the lower the distance between similar terms, the better the contextuality. Figure \ref{fig:moreInstrinsic} shows additional intrinsic evaluation examples.

\newpage
\subsection*{Additional examples of image captioning - IMAGINATOR vs. Gold Caption vs. OSCAR (SoTA)}
\bigskip
\bigskip

\begin{figure}[!h]
\centering
\begin{minipage}{.3\textwidth}
  \centering
\includegraphics[width=0.7\textwidth]{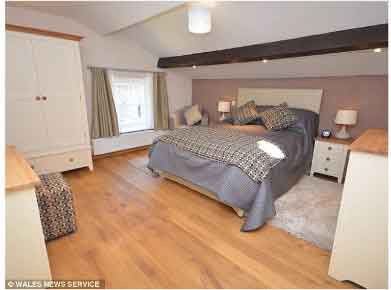}
    \subcaption{
    \textbf{IMAGINATOR}: A bedroom with bedspreads, pillows, and a nightstand.\\
    \textbf{Gold Caption}: the - bedroom stone cottage can sleep people.\\
    \textbf{OSCAR}: A bedroom with a bed, dresser, and nightstand.}
\end{minipage}%
\hspace{0.5cm}
\begin{minipage}{.3\textwidth}
  \centering
\includegraphics[width=0.7\textwidth]{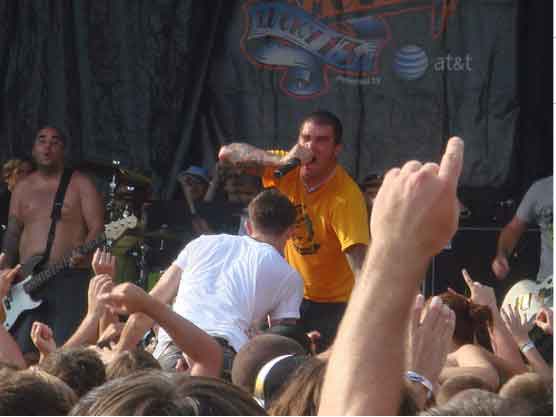}
    \subcaption{
    \textbf{IMAGINATOR}: A group of people are singing and clapping while a group of musicians are performing.\\
    \textbf{Gold Caption}: A band is playing in front of an audience and the singer is wearing an orange shirt.\\
    \textbf{OSCAR}: A man holding a baseball bat in front of a crowd.\\
    }
\end{minipage}%
\hspace{0.5cm}
\begin{minipage}{.3\textwidth}
  \centering
\includegraphics[width=0.7\textwidth]{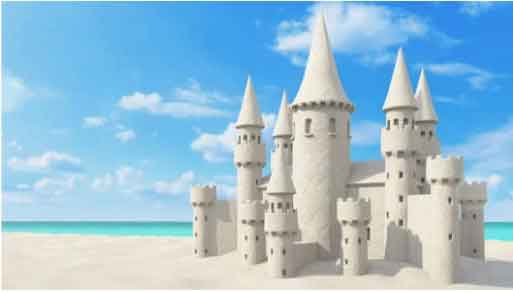}
    \subcaption{
    \textbf{IMAGINATOR}: Photograph of a tall tower with
steeples.\\
    \textbf{Gold Caption}: sandcastle beach on bright sky.\\
    \textbf{OSCAR}: A castle made of sand with a clock tower
in the background.}
\end{minipage}%

\bigskip
\bigskip

\begin{minipage}{.3\textwidth}
  \centering
\includegraphics[width=0.7\textwidth]{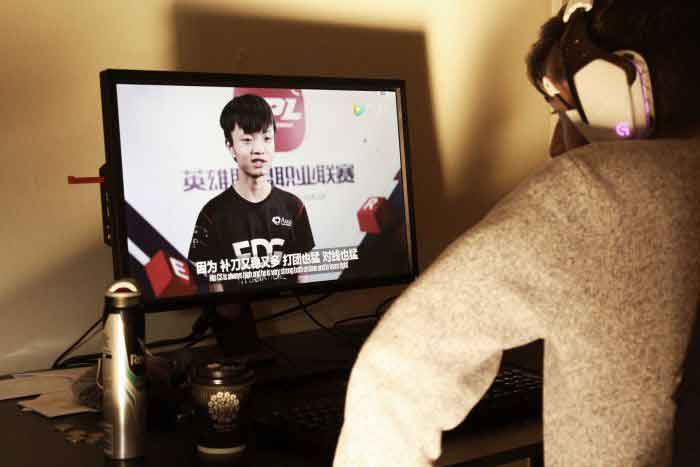}
    \subcaption{
    \textbf{IMAGINATOR}: A photo of a console.\\
    \textbf{Gold Caption}: The player staring intently at a
computer screen.\\
    \textbf{OSCAR}: A man sitting in front of a flat screen TV.}
\end{minipage}%
\hspace{0.5cm}
\begin{minipage}{.3\textwidth}
  \centering
\includegraphics[width=0.7\textwidth]{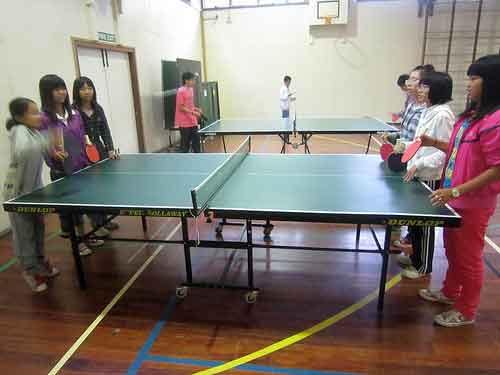}
    \subcaption{
    \textbf{IMAGINATOR}: A group of people playing
ping-pong together.\\
    \textbf{Gold Caption}: Young girls line up across each
other and a ping-pong table in a gymnasium while
a few boys plan on a table further back.\\
    \textbf{OSCAR}: A group of children playing a game of
ping pong.}
\end{minipage}%
\hspace{0.5cm}
\begin{minipage}{.3\textwidth}
  \centering
\includegraphics[width=0.7\textwidth]{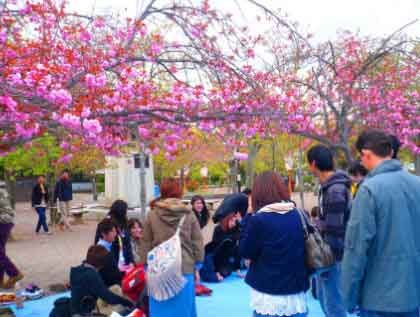}
    \subcaption{
    \textbf{IMAGINATOR}: "girls" and "boys" at a venue.\\
    \textbf{Gold Caption}:  party in the park under cherry
blossoms.\\
    \textbf{OSCAR}: A group of people sitting around a park
with pink flowers.}
\end{minipage}%

\bigskip
\bigskip
\bigskip

\begin{minipage}{.3\textwidth}
  \centering
\includegraphics[width=0.7\textwidth]{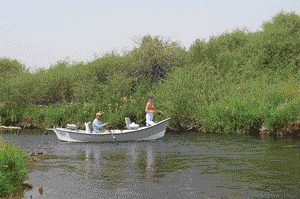}
    \subcaption{
    \textbf{IMAGINATOR}: A young woman and
man rowing a boat.\\
    \textbf{Gold Caption}: A man and woman are
on a gray and white rowboat.\\
    \textbf{OSCAR}: Group of people on a
small boat in the water.}
\end{minipage}%
\hspace{0.5cm}
\begin{minipage}{.3\textwidth}
  \centering
\includegraphics[width=0.7\textwidth]{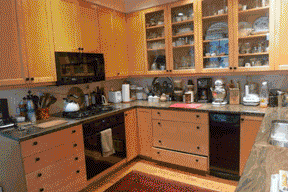}
    \subcaption{
    \textbf{IMAGINATOR}: A kitchen with cabinets, cabinets, and a dishwasher.\\
    \textbf{Gold Caption}: A kitchen with wooden
cabinets and black appliances \\
    \textbf{OSCAR}: A kitchen with a sink,
dishwasher, stove and refrigerator.}
\end{minipage}%
\hspace{0.5cm}
\begin{minipage}{.3\textwidth}
  \centering
\includegraphics[width=0.7\textwidth]{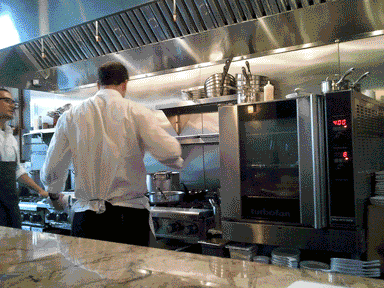}
    \subcaption{
    \textbf{IMAGINATOR}: Chefs cooking with a stover and other cookware in a laboratory.\\
    \textbf{Gold Caption}:  Two chefs in a restaurant kitchen
preparing food .\\
    \textbf{OSCAR}: Two men in a commercial kitchen
preparing food.}
\end{minipage}%

\bigskip
\caption{Examples of some image captioning outputs generated by IMAGINATOR along with the original caption and the caption generated by OSCAR \cite{li2020oscar} for each respective image}

\label{fig:moreCaptionexamples}
\end{figure}

\newpage
\section*{Examples of image retrieval - IMAGINATOR vs. SoTA: (i) ALBEF \cite{albef}, and (ii) XVLM \cite{xvlm}}

\begin{figure*}[htp!]
\centering
\begin{subfigure}{0.35\textwidth}
  \centering
   \includegraphics[width=0.7\textwidth]{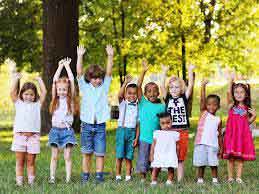}
    \caption{ rank 1 image
    ALBEF \cite{albef}
    }
\end{subfigure}%
\begin{subfigure}{0.33\textwidth}
  \centering
   \includegraphics[width=0.7\textwidth]{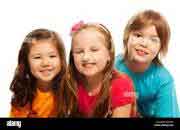}
    \caption{ rank 1 image 
    XVLM \cite{xvlm}
    }
\end{subfigure}%
\begin{subfigure}{0.34\textwidth}
  \centering
   \includegraphics[width=0.7\textwidth]{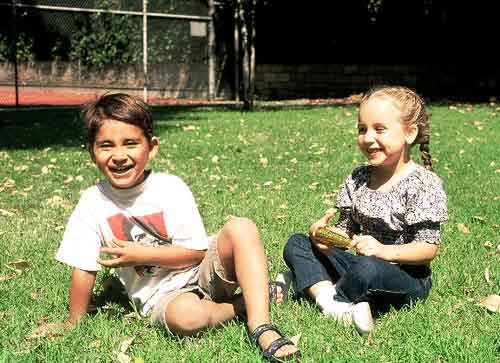}
    \caption{ rank 1 image
    BERT$_{IMAGINATOR}$
    }
\end{subfigure}%
\caption{Image retrieved for the query: \textit{"Two little children, one boy and one girl laughing"}.}
\label{fig:IR_Captions_1}
\end{figure*}

\begin{figure*}[htp!]
\centering
\begin{subfigure}{0.35\textwidth}
  \centering
   \includegraphics[width=0.7\textwidth]{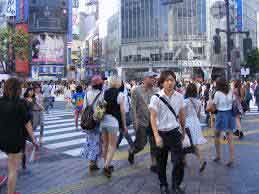}
    \caption{ rank 1 image
    ALBEF \cite{albef}
    }
\end{subfigure}%
\begin{subfigure}{0.33\textwidth}
  \centering
   \includegraphics[width=0.7\textwidth]{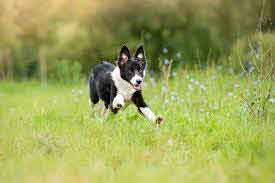}
    \caption{ rank 1 image 
    XVLM \cite{xvlm}
    }
\end{subfigure}%
\begin{subfigure}{0.34\textwidth}
  \centering
   \includegraphics[width=0.7\textwidth]{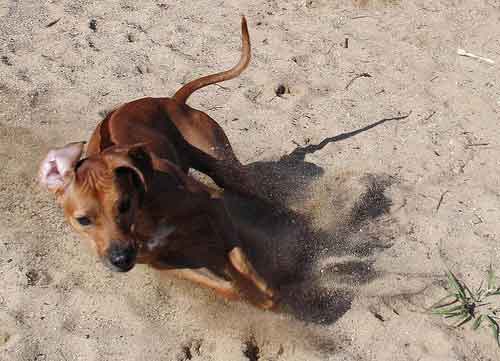}
    \caption{ rank 1 image
    BERT$_{IMAGINATOR}$
    }
\end{subfigure}%
\caption{Image retrieved for the query: \textit{"A dog is running in the sand"}.}
\label{fig:IR_Captions_2}
\end{figure*}

\begin{figure*}[htp!]
\centering
\begin{subfigure}{0.35\textwidth}
  \centering
   \includegraphics[width=0.7\textwidth]{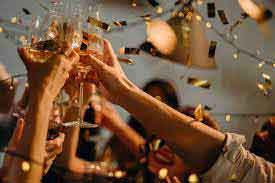}
    \caption{ rank 1 image
    ALBEF \cite{albef}
    }
\end{subfigure}%
\begin{subfigure}{0.33\textwidth}
  \centering
   \includegraphics[width=0.7\textwidth]{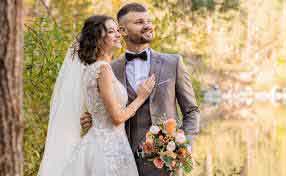}
    \caption{ rank 1 image 
    XVLM \cite{xvlm}
    }
\end{subfigure}%
\begin{subfigure}{0.34\textwidth}
  \centering
   \includegraphics[width=0.7\textwidth]{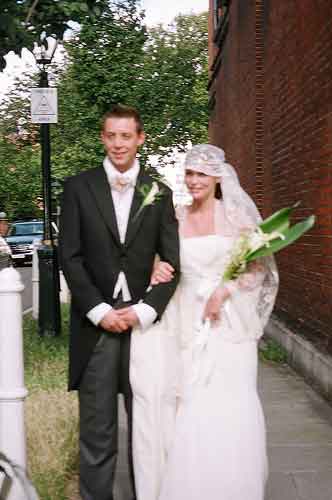}
    \caption{ rank 1 image
    BERT$_{IMAGINATOR}$
    }
\end{subfigure}%
\caption{Image retrieved for the query: \textit{"Bride and groom walking
side by side".}}
\label{fig:IR_Captions_3}
\end{figure*}

\begin{figure*}[htp!]
\centering
\begin{subfigure}{0.35\textwidth}
  \centering
   \includegraphics[width=0.7\textwidth]{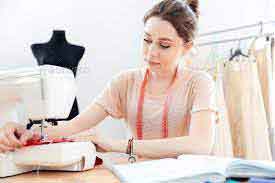}
    \caption{ rank 1 image
    ALBEF \cite{albef}
    }
\end{subfigure}%
\begin{subfigure}{0.33\textwidth}
  \centering
   \includegraphics[width=0.7\textwidth]{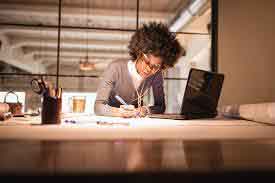}
    \caption{ rank 1 image 
    XVLM \cite{xvlm}
    }
\end{subfigure}%
\begin{subfigure}{0.34\textwidth}
  \centering
   \includegraphics[width=0.7\textwidth]{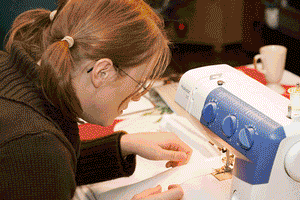}
    \caption{ rank 1 image
    BERT$_{IMAGINATOR}$
    }
\end{subfigure}%
\caption{Image retrieved for the query: \textit{"Redhead woman in pig-
tails and glasses sewing on a
sewing machine".}}
\label{fig:IR_Captions_4}
\end{figure*}

\begin{figure*}[htp!]
\centering
\begin{subfigure}{0.35\textwidth}
  \centering
   \includegraphics[width=0.7\textwidth]{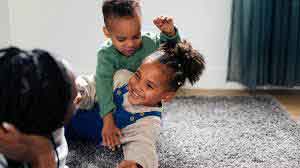}
    \caption{ rank 1 image
    ALBEF \cite{albef}
    }
\end{subfigure}%
\begin{subfigure}{0.33\textwidth}
  \centering
   \includegraphics[width=0.7\textwidth]{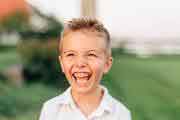}
    \caption{ rank 1 image 
    XVLM \cite{xvlm}
    }
\end{subfigure}%
\begin{subfigure}{0.34\textwidth}
  \centering
   \includegraphics[width=0.7\textwidth]{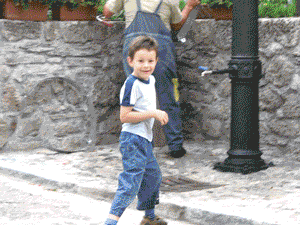}
    \caption{ rank 1 image
    BERT$_{IMAGINATOR}$
    }
\end{subfigure}%
\caption{Image retrieved for the query: \textit{"Smiling boy in white shirt and blue jeans in front of rock wall with man in overalls behind him".}}
\label{fig:IR_Captions_5}
\end{figure*}

\begin{figure*}[htp!]
\centering
\begin{subfigure}{0.35\textwidth}
  \centering
   \includegraphics[width=0.7\textwidth]{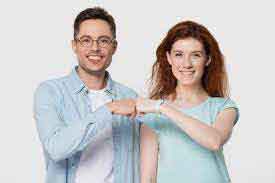}
    \caption{ rank 1 image
    ALBEF \cite{albef}
    }
\end{subfigure}%
\begin{subfigure}{0.33\textwidth}
  \centering
   \includegraphics[width=0.7\textwidth]{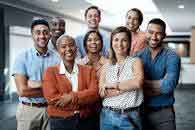}
    \caption{ rank 1 image 
    XVLM \cite{xvlm}
    }
\end{subfigure}%
\begin{subfigure}{0.34\textwidth}
  \centering
   \includegraphics[width=0.7\textwidth]{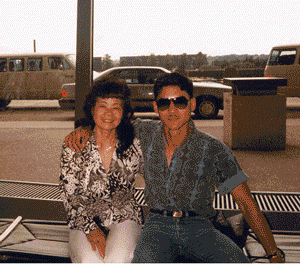}
    \caption{ rank 1 image
    BERT$_{IMAGINATOR}$
    }
\end{subfigure}%
\caption{Image retrieved for the query: \textit{"Two Asian or Spanish people, a woman and a man,
sitting together in front of a
glass window as cars pass".}}
\label{fig:IR_Captions_6}
\end{figure*}

\begin{figure*}[htp!]
\centering
\begin{subfigure}{0.35\textwidth}
  \centering
   \includegraphics[width=0.7\textwidth]{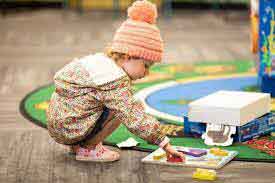}
    \caption{ rank 1 image
    ALBEF \cite{albef}
    }
\end{subfigure}%
\begin{subfigure}{0.33\textwidth}
  \centering
   \includegraphics[width=0.7\textwidth]{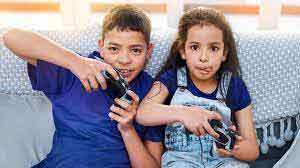}
    \caption{ rank 1 image 
    XVLM \cite{xvlm}
    }
\end{subfigure}%
\begin{subfigure}{0.34\textwidth}
  \centering
   \includegraphics[width=0.7\textwidth]{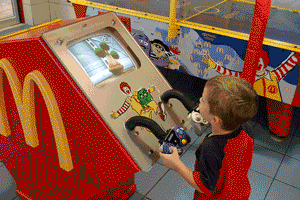}
    \caption{ rank 1 image
    BERT$_{IMAGINATOR}$
    }
\end{subfigure}%
\caption{Image retrieved for the query: \textit{"A little boy plays with
a Nintendo GameCube controller inside a McDonald’s".}}
\label{fig:IR_Captions_7}
\end{figure*}

\begin{figure*}[htp!]
\centering
\begin{subfigure}{0.35\textwidth}
  \centering
   \includegraphics[width=0.7\textwidth]{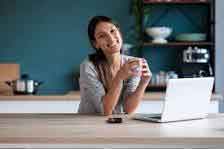}
    \caption{ rank 1 image
    ALBEF \cite{albef}
    }
\end{subfigure}%
\begin{subfigure}{0.33\textwidth}
  \centering
   \includegraphics[width=0.7\textwidth]{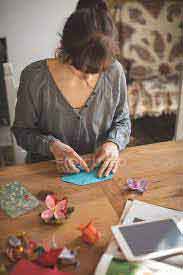}
    \caption{ rank 1 image 
    XVLM \cite{xvlm}
    }
\end{subfigure}%
\begin{subfigure}{0.34\textwidth}
  \centering
   \includegraphics[width=0.7\textwidth]{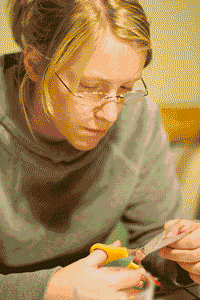}
    \caption{ rank 1 image
    BERT$_{IMAGINATOR}$
    }
\end{subfigure}%
\caption{Image retrieved for the query: \textit{"A blonde woman wearing
glasses and a gray sweatshirt is cutting something with scissors".}}
\label{fig:IR_Captions_8}
\end{figure*}

\begin{figure*}[htp!]
\centering
\begin{subfigure}{0.35\textwidth}
  \centering
   \includegraphics[width=0.7\textwidth]{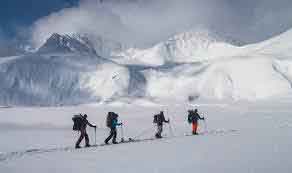}
    \caption{ rank 1 image
    ALBEF \cite{albef}
    }
\end{subfigure}%
\begin{subfigure}{0.33\textwidth}
  \centering
   \includegraphics[width=0.7\textwidth]{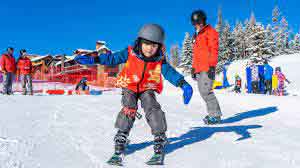}
    \caption{ rank 1 image 
    XVLM \cite{xvlm}
    }
\end{subfigure}%
\begin{subfigure}{0.34\textwidth}
  \centering
   \includegraphics[width=0.7\textwidth]{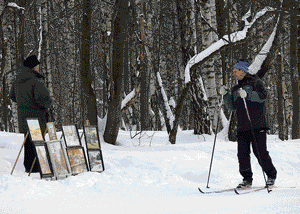}
    \caption{ rank 1 image
    BERT$_{IMAGINATOR}$
    }
\end{subfigure}%
\caption{Image retrieved for the query: \textit{"A person wearing skis
looking at framed pictures set
up in the snow".}}
\label{fig:IR_Captions_9}
\end{figure*}

\begin{figure*}[htp!]
\centering
\begin{subfigure}{0.35\textwidth}
  \centering
   \includegraphics[width=0.7\textwidth]{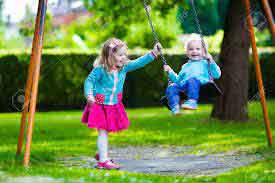}
    \caption{ rank 1 image
    ALBEF \cite{albef}
    }
\end{subfigure}%
\begin{subfigure}{0.33\textwidth}
  \centering
   \includegraphics[width=0.7\textwidth]{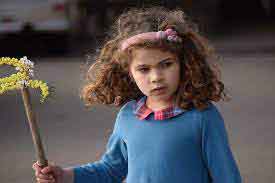}
    \caption{ rank 1 image 
    XVLM \cite{xvlm}
    }
\end{subfigure}%
\begin{subfigure}{0.34\textwidth}
  \centering
   \includegraphics[width=0.7\textwidth]{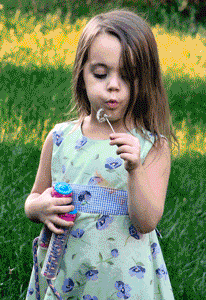}
    \caption{ rank 1 image
    BERT$_{IMAGINATOR}$
    }
\end{subfigure}%
\caption{Image retrieved for the query: \textit{"A very young girl playing
with a bubble-blowing wand
, holding a bottle of bubble
solution and walking through
a park or field".}}
\label{fig:IR_Captions_10}
\end{figure*}
\begin{figure*}[htp!]
\centering
\begin{subfigure}{0.35\textwidth}
  \centering
   \includegraphics[width=0.7\textwidth]{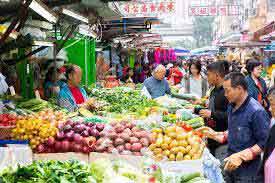}
    \caption{ rank 1 image
    ALBEF \cite{albef}
    }
\end{subfigure}%
\begin{subfigure}{0.33\textwidth}
  \centering
   \includegraphics[width=0.7\textwidth]{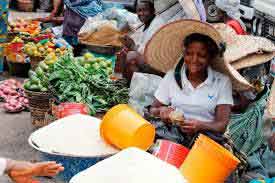}
    \caption{ rank 1 image 
    XVLM \cite{xvlm}
    }
\end{subfigure}%
\begin{subfigure}{0.34\textwidth}
  \centering
   \includegraphics[width=0.7\textwidth]{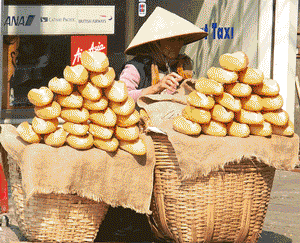}
    \caption{ rank 1 image
    BERT$_{IMAGINATOR}$
    }
\end{subfigure}%
\caption{Image retrieved for the query: \textit{"A woman in an outdoor
marketplace, wearing a large
cone-shaped hat, standing behind two large baskets containing loaves of bread".}}
\label{fig:IR_Captions_11}
\end{figure*}

\clearpage
\subsection*{Image2Tweet examples - Gold vs. 5 ensemble SoTA \cite{lu2018entity} vs. $BERT_{IMAGINATOR}$ }

\begin{figure}[!h]
\centering
\begin{minipage}{.3\textwidth}
  \centering
\includegraphics[width=0.6\textwidth]{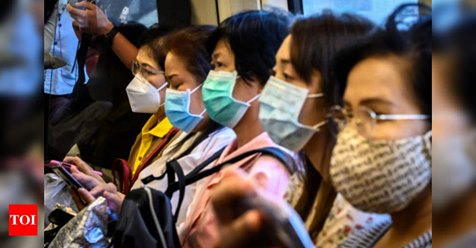}
\subcaption{
\fontsize{8}{10}\selectfont
    \textbf{Gold Caption}: Should you wear a mask to protect yourself from \#coronavirus? \#Coronavirus \#COVID19 \\
    \textbf{5 ensemble \cite{luo2018discriminability}}: a group of surgeons prepare for surgery.\\
    \textbf{IMAGINATOR}: people wearing masks during the pandemic.
}
\end{minipage}%
\hspace{0.5cm}
\begin{minipage}{.3\textwidth}
  \centering
\includegraphics[width=0.6\textwidth]{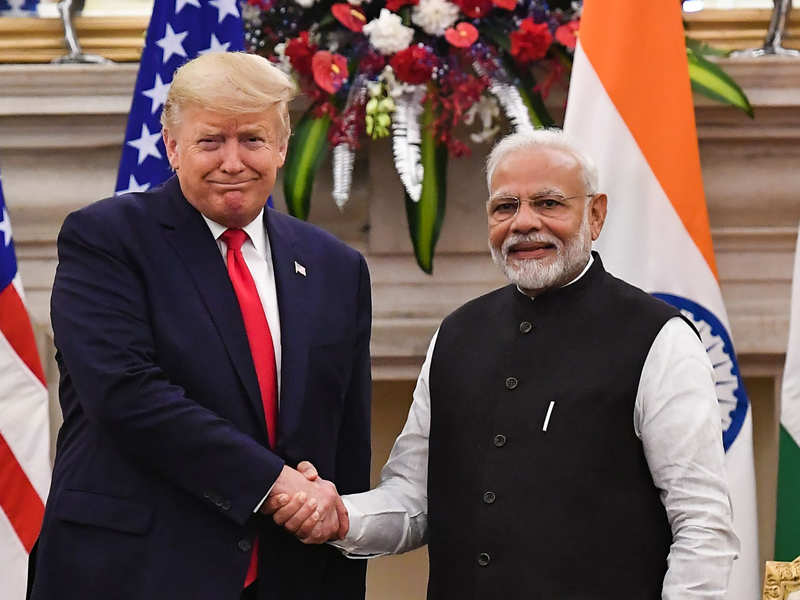}
    \subcaption{
\fontsize{8}{10}\selectfont    
    \textbf{Gold Caption}: Donald Trump's India visit will be beneficial for both the countries.\\
    \textbf{5 ensemble \cite{luo2018discriminability}}: politician shakes hands with politician during a bilateral meeting.\\
    \textbf{IMAGINATOR}: Two men are handshaking with an Indian flag in the background.
}
\end{minipage}%
\hspace{0.5cm}
\begin{minipage}{.3\textwidth}
  \centering
\includegraphics[width=0.6\textwidth]{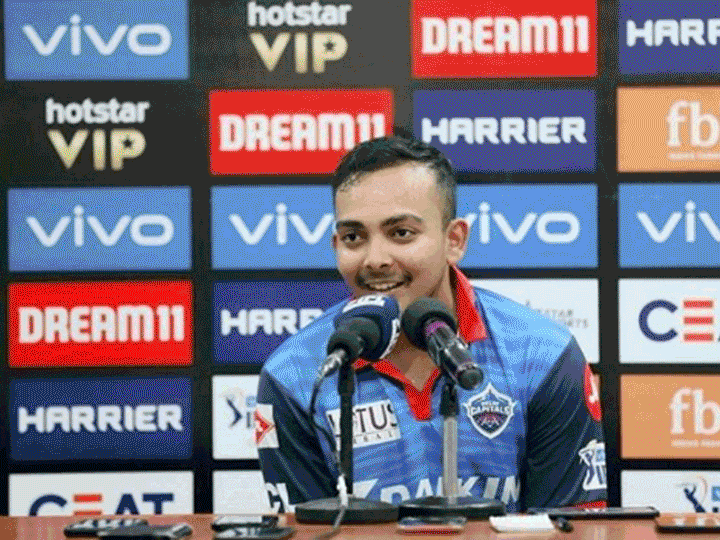}
    \subcaption{
\fontsize{8}{10}\selectfont    
    \textbf{Gold Caption}: I am here to play cricket not gimmick - @PrithviShaw to press.\\
    \textbf{5 ensemble \cite{luo2018discriminability}}: cricket player during a press conference.\\
    \textbf{IMAGINATOR}: A man in a press conference.
}
\end{minipage}%

\begin{minipage}{.3\textwidth}
  \centering
\includegraphics[width=0.6\textwidth]{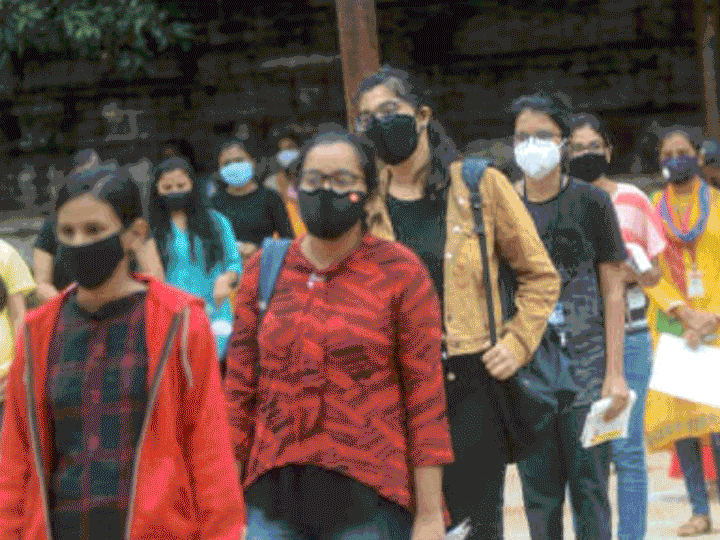}
    \subcaption{
\fontsize{8}{10}\selectfont    
   \textbf{Gold Caption}: JEE (Main) begins today - students are following protocols - queue, social distancing, masks.\\
    \textbf{5 ensemble \cite{luo2018discriminability}}: students wearing face masks during a protest.
    \\
    \textbf{IMAGINATOR}: young girls wearing masks in a queue.
}
\end{minipage}%
\hspace{0.5cm}
\begin{minipage}{.3\textwidth}
  \centering
\includegraphics[width=0.6\textwidth]{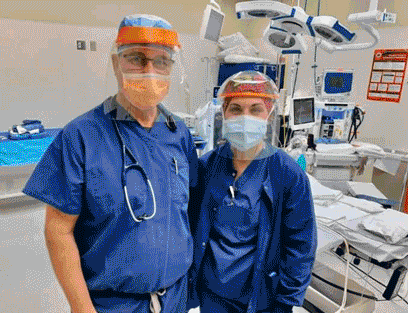}
    \subcaption{
\fontsize{8}{10}\selectfont    
    \textbf{Gold Caption}: Country needs so many doctors than politicians - pandemic realization.
    \\
    \textbf{5 ensemble \cite{luo2018discriminability}}: person, left, and person, right, are both members of the team.
    \\
    \textbf{IMAGINATOR}: Two doctors with face shields.
}
\end{minipage}%
\hspace{0.5cm}
\begin{minipage}{.3\textwidth}
  \centering
\includegraphics[width=0.6\textwidth]{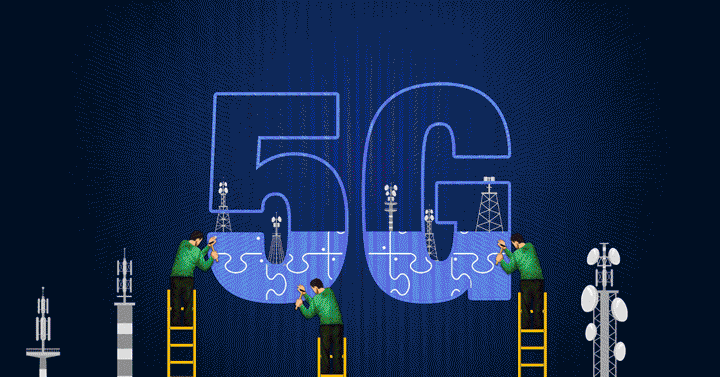}
    \subcaption{
    \fontsize{8}{10}\selectfont
   \textbf{Gold Caption}: 5G tech is picking up pace and expectations are high, but rollout is still years away in India.\\
   \textbf{5 ensemble \cite{luo2018discriminability}}: the logo on a background of a blue sky with clouds.
   \\
    \textbf{IMAGINATOR}: 5G logo.
}
\end{minipage}%
\\
\begin{minipage}{.3\textwidth}
  \centering
\includegraphics[width=0.6\textwidth]{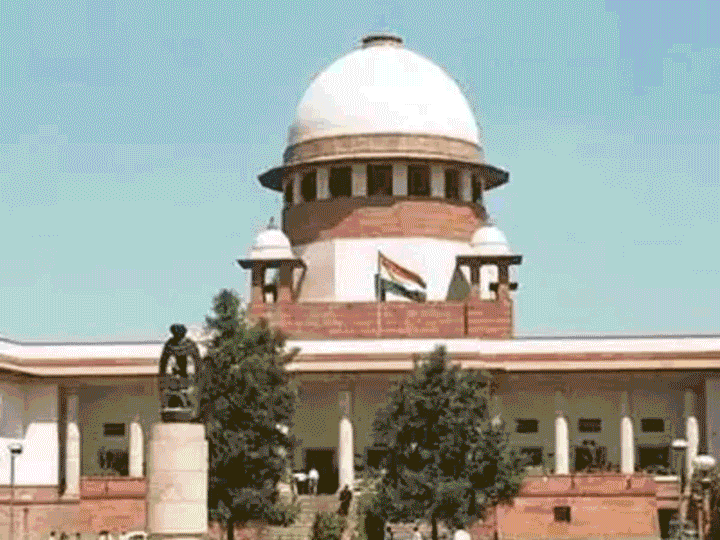}
    \subcaption{
\fontsize{8}{10}\selectfont    
    \textbf{Gold Caption}: SC refuses to entertain plea against Madras HC order on Patanjali's use of 'Coronil'.\\
    \textbf{5 ensemble \cite{luo2018discriminability}}: a gothic buildiing.
    \\
    \textbf{IMAGINATOR}: supreme court of India building.
}
\end{minipage}%
\hspace{0.5cm}
\begin{minipage}{.3\textwidth}
  \centering
\includegraphics[width=0.6\textwidth]{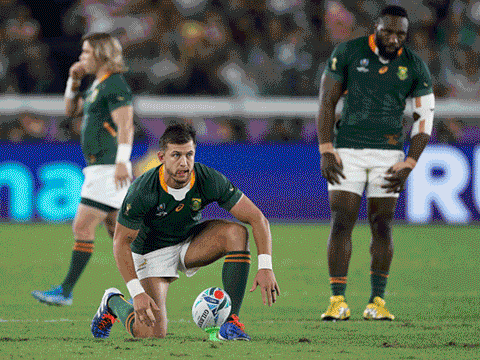}
    \subcaption{
\fontsize{8}{10}\selectfont    
  \textbf{Gold Caption}:  No rugby for world champion as South Africa maintains ban.\\
    \textbf{5 ensemble \cite{luo2018discriminability}}: rugby player looks dejected after defeat
    \\
    \textbf{IMAGINATOR}: A scene of a rugby match with three players visible.
}
\end{minipage}%
\hspace{0.5cm}
\begin{minipage}{.3\textwidth}
  \centering
\includegraphics[width=0.6\textwidth]{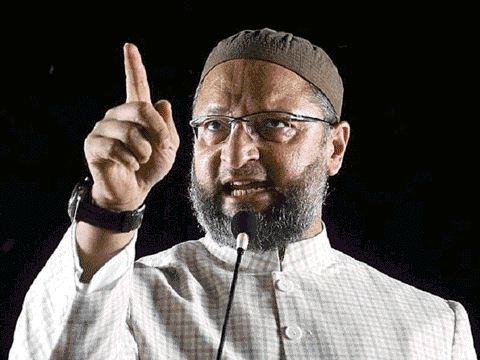}
    \subcaption{
 \textbf{Gold Caption}: I love India, but Indians don't like me.\\
    \textbf{5 ensemble \cite{luo2018discriminability}}: politician addresses a crowd of supporters.
    \\
    \textbf{IMAGINATOR}: An angry politician delivering a speech.
}
\end{minipage}%
\\\begin{minipage}{.3\textwidth}
  \centering
\includegraphics[width=0.6\textwidth]{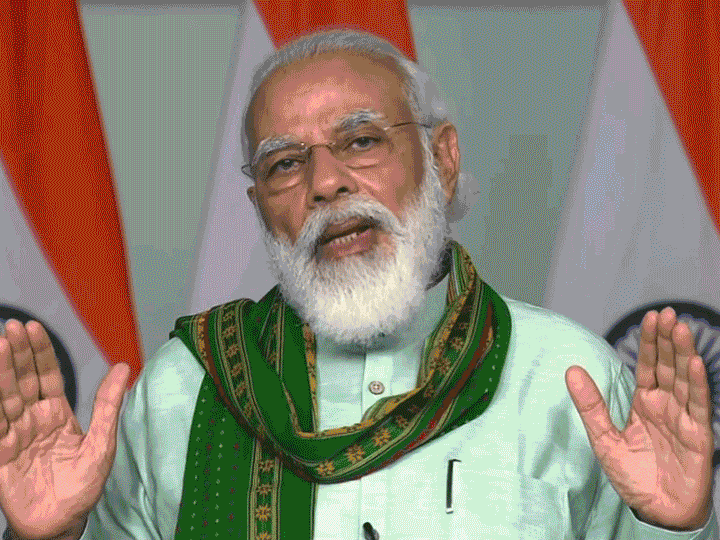}
    \subcaption{
\fontsize{8}{10}\selectfont    
 \textbf{Gold Caption}: Indian prime minister addressing to the nation in his own man ki baat.\\
    \textbf{5 ensemble \cite{luo2018discriminability}}: politician making a speech at a function.
    \\
    \textbf{IMAGINATOR}: Modi is delivering a speech on camera.
}
\end{minipage}%
\hspace{0.5cm}
\begin{minipage}{.3\textwidth}
  \centering
\includegraphics[width=0.6\textwidth]{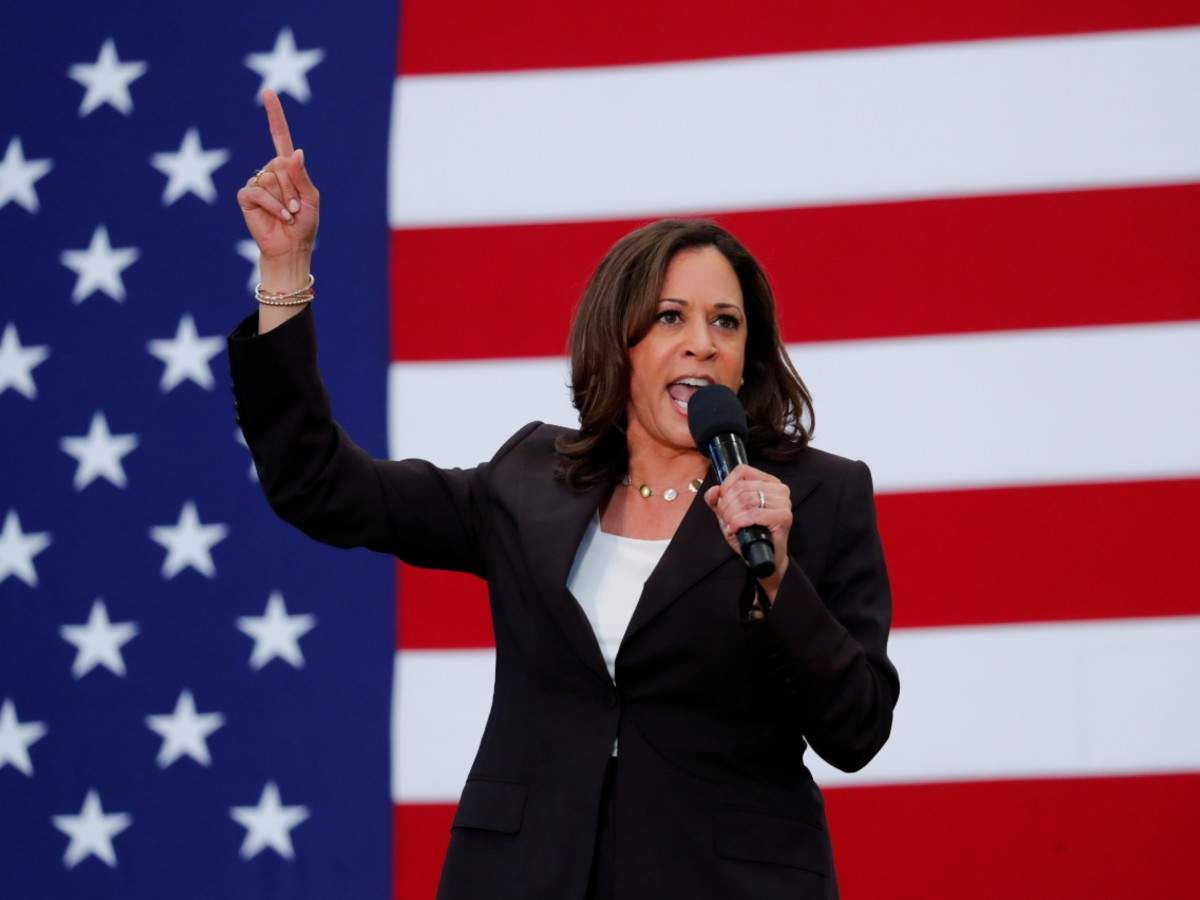}
    \subcaption{
\fontsize{8}{10}\selectfont    
 \textbf{Gold Caption}: Kamala Harris bringing energy, dollars and more to Joe Biden's campaign.\\
    \textbf{5 ensemble \cite{luo2018discriminability}}: politician gives a speech during the second day.
    \\
    \textbf{IMAGINATOR}: Harris making promises.}

\end{minipage}%
\hspace{0.5cm}
\begin{minipage}{.3\textwidth}
  \centering
\includegraphics[width=0.6\textwidth]{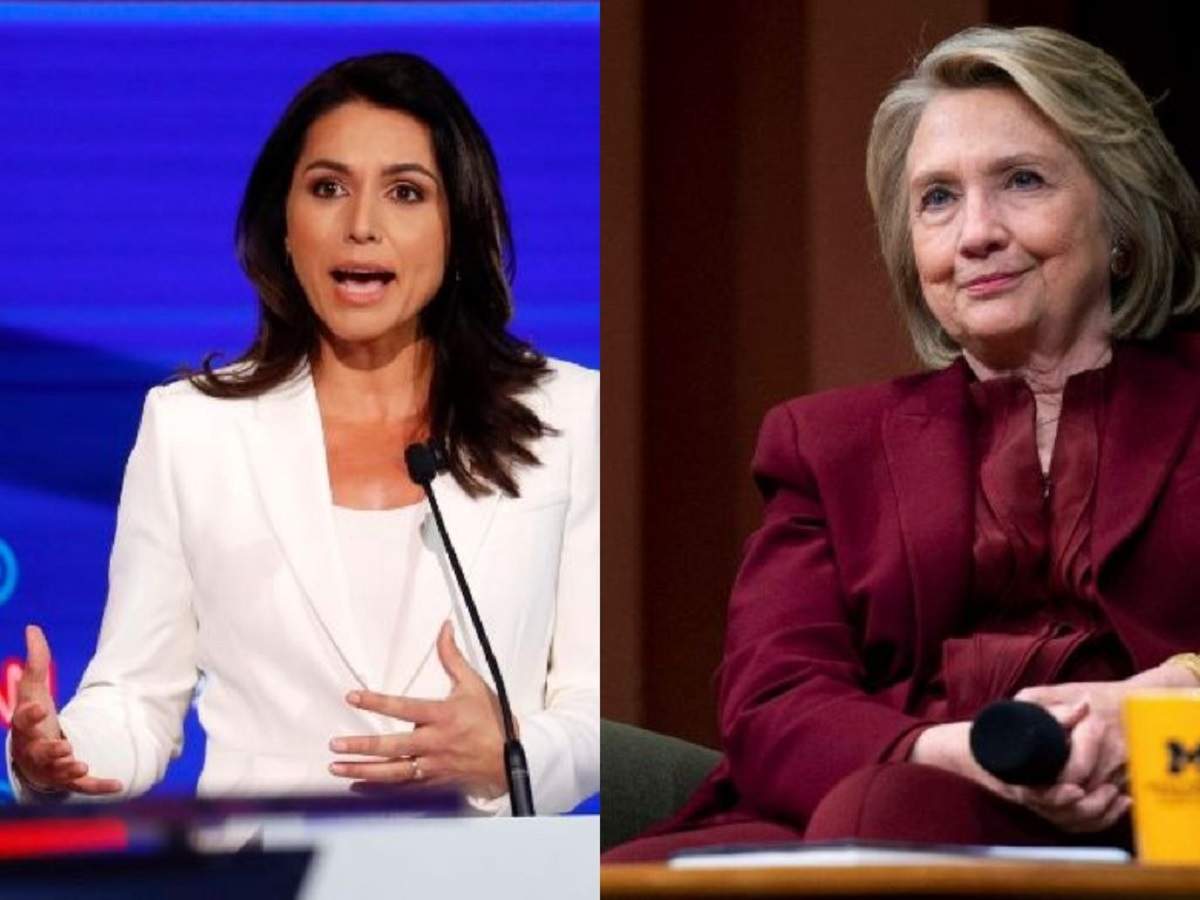}
    \subcaption{
\fontsize{8}{10}\selectfont    
 \textbf{Gold Caption}:   US Presidential election: Hillary-Tulsi spat scorches Democratic Party.\\
    \textbf{5 ensemble \cite{luo2018discriminability}}: Two politicians are debating.
    \\
    \textbf{IMAGINATOR}: Hillary Clinton and another woman in white dress.
}
\end{minipage}%
\\
\caption{Examples of Image2Tweet task - gold vs IMAGINATOR vs 5 ensemble SoTA \cite{li2020oscar}}
\label{fig:moreimg2tweet_examples}
\end{figure}

Image2tweet is a particularly hard problem to solve. It can involve social engineering, web information scraping, face recognition, etc. The results in table \ref{table:img2tweet} show the current status of the problem and it needs substantial research work to develop a solution. Figure \ref{fig:moreimg2tweet_examples} shows some Image2Tweet examples.

%% file: 0_main.bbl
\begin{thebibliography}{48}
\expandafter\ifx\csname natexlab\endcsname\relax\def\natexlab#1{#1}\fi

\bibitem[{Andrew et~al.(2013)Andrew, Arora, Bilmes, and
  Livescu}]{andrew2013deep}
Galen Andrew, Raman Arora, Jeff Bilmes, and Karen Livescu. 2013.
\newblock Deep canonical correlation analysis.
\newblock In \emph{International conference on machine learning}, pages
  1247--1255. PMLR.

\bibitem[{Artetxe et~al.(2018)Artetxe, Labaka, and
  Agirre}]{artetxe2018generalizing}
Mikel Artetxe, Gorka Labaka, and Eneko Agirre. 2018.
\newblock Generalizing and improving bilingual word embedding mappings with a
  multi-step framework of linear transformations.
\newblock In \emph{Proceedings of the AAAI Conference on Artificial
  Intelligence}, volume~32.

\bibitem[{Baroni and Lenci(2011)}]{baroni-lenci-2011-blessed}
Marco Baroni and Alessandro Lenci. 2011.
\newblock \href {https://aclanthology.org/W11-2501} {How we {BLESS}ed
  distributional semantic evaluation}.
\newblock In \emph{Proceedings of the {GEMS} 2011 Workshop on {GE}ometrical
  Models of Natural Language Semantics}, pages 1--10, Edinburgh, UK.
  Association for Computational Linguistics.

\bibitem[{Bhatia(2021)}]{bhatia}
Gagan Bhatia. 2021.
\newblock \href {https://github.com/gagan3012/keytotext} {keytotext}.

\bibitem[{Bruni et~al.(2014)Bruni, Tran, and Baroni}]{10.5555/2655713.2655714}
Elia Bruni, Nam~Khanh Tran, and Marco Baroni. 2014.
\newblock Multimodal distributional semantics.
\newblock \emph{J. Artif. Int. Res.}, 49(1):1–47.

\bibitem[{Chen et~al.(2020)Chen, Li, Yu, El~Kholy, Ahmed, Gan, Cheng, and
  Liu}]{chen2020uniter}
Yen-Chun Chen, Linjie Li, Licheng Yu, Ahmed El~Kholy, Faisal Ahmed, Zhe Gan,
  Yu~Cheng, and Jingjing Liu. 2020.
\newblock Uniter: Universal image-text representation learning.
\newblock In \emph{European conference on computer vision}, pages 104--120.
  Springer.

\bibitem[{Devlin et~al.(2018)Devlin, Chang, Lee, and
  Toutanova}]{devlin2018bert}
Jacob Devlin, Ming-Wei Chang, Kenton Lee, and Kristina Toutanova. 2018.
\newblock Bert: Pre-training of deep bidirectional transformers for language
  understanding.
\newblock \emph{arXiv preprint arXiv:1810.04805}.

\bibitem[{Finkelstein et~al.(2001)Finkelstein, Gabrilovich, Matias, Rivlin,
  Solan, Wolfman, and Ruppin}]{finkelstein-2001-placing}
Lev Finkelstein, Evgeniy Gabrilovich, Yossi Matias, Ehud Rivlin, Zach Solan,
  Gadi Wolfman, and Eytan Ruppin. 2001.
\newblock \href {https://doi.org/10.1145/371920.372094} {Placing search in
  context: the concept revisited}.
\newblock In \emph{Proceedings of the Tenth International World Wide Web
  Conference, {WWW} 10, Hong Kong, China, May 1-5, 2001}, pages 406--414.
  {ACM}.

\bibitem[{Firth(1957)}]{firth1957}
John~Rupert Firth. 1957.
\newblock \emph{"A synopsis of linguistic theory 1930-1955."}.
\newblock Oxford University Press.

\bibitem[{Gunti et~al.(2022)Gunti, Ramamoorthy, Patwa, and
  Das}]{Gunti-Ramamoorthy-Patwa-Das-2022}
Nethra Gunti, Sathyanarayanan Ramamoorthy, Parth Patwa, and Amitava Das. 2022.
\newblock \href {https://doi.org/10.1609/aaai.v36i11.21616} {Memotion analysis
  through the lens of joint embedding (student abstract)}.
\newblock \emph{Proceedings of the AAAI Conference on Artificial Intelligence},
  36(11):12959--12960.

\bibitem[{Hadsell et~al.(2006)Hadsell, Chopra, and
  LeCun}]{hadsell2006dimensionality}
Raia Hadsell, Sumit Chopra, and Yann LeCun. 2006.
\newblock Dimensionality reduction by learning an invariant mapping.
\newblock In \emph{2006 IEEE Computer Society Conference on Computer Vision and
  Pattern Recognition (CVPR'06)}, volume~2, pages 1735--1742. IEEE.

\bibitem[{Halawi et~al.(2012)Halawi, Dror, Gabrilovich, and
  Koren}]{10.1145/2339530.2339751}
Guy Halawi, Gideon Dror, Evgeniy Gabrilovich, and Yehuda Koren. 2012.
\newblock \href {https://doi.org/10.1145/2339530.2339751} {Large-scale learning
  of word relatedness with constraints}.
\newblock In \emph{Proceedings of the 18th ACM SIGKDD International Conference
  on Knowledge Discovery and Data Mining}, KDD '12, page 1406–1414, New York,
  NY, USA. Association for Computing Machinery.

\bibitem[{Hardoon et~al.(2004)Hardoon, Szedmak, and
  Shawe-Taylor}]{hardoon2004canonical}
David~R Hardoon, Sandor Szedmak, and John Shawe-Taylor. 2004.
\newblock Canonical correlation analysis: An overview with application to
  learning methods.
\newblock \emph{Neural computation}, 16(12):2639--2664.

\bibitem[{Harris(1954)}]{distributional}
Zellig Harris. 1954.
\newblock \href {https://doi.org/10.1007/978-94-009-8467-7_1} {{Distributional
  Structure}}.
\newblock \emph{Word}, 10:146--162.

\bibitem[{Hill et~al.(2015)Hill, Reichart, and
  Korhonen}]{hill-etal-2015-simlex}
Felix Hill, Roi Reichart, and Anna Korhonen. 2015.
\newblock \href {https://doi.org/10.1162/COLI_a_00237} {{S}im{L}ex-999:
  Evaluating semantic models with (genuine) similarity estimation}.
\newblock \emph{Computational Linguistics}, 41(4):665--695.

\bibitem[{Jastrzebski et~al.(2017)Jastrzebski, Lesniak, and
  Czarnecki}]{DBLP:journals/corr/JastrzebskiLC17}
Stanislaw Jastrzebski, Damian Lesniak, and Wojciech~Marian Czarnecki. 2017.
\newblock How to evaluate word embeddings? on importance of data efficiency and
  simple supervised tasks.
\newblock \emph{CoRR}, abs/1702.02170.

\bibitem[{Jha et~al.(2021)Jha, Kaki, Kolla, Bhagat, Patwa, Das, and
  Pal}]{jha-etal-2021-image2tweet}
Rishabh Jha, Varshith Kaki, Varuna Kolla, Shubham Bhagat, Parth Patwa, Amitava
  Das, and Santanu Pal. 2021.
\newblock \href {https://aclanthology.org/2021.icon-main.84} {Image2tweet:
  Datasets in {H}indi and {E}nglish for generating tweets from images}.
\newblock In \emph{Proceedings of the 18th International Conference on Natural
  Language Processing (ICON)}, pages 670--676, National Institute of Technology
  Silchar, Silchar, India. NLP Association of India (NLPAI).

\bibitem[{Jia et~al.(2021)Jia, Yang, Xia, Chen, Parekh, Pham, Le, Sung, Li, and
  Duerig}]{jia2021scaling}
Chao Jia, Yinfei Yang, Ye~Xia, Yi-Ting Chen, Zarana Parekh, Hieu Pham, Quoc Le,
  Yun-Hsuan Sung, Zhen Li, and Tom Duerig. 2021.
\newblock Scaling up visual and vision-language representation learning with
  noisy text supervision.
\newblock In \emph{International Conference on Machine Learning}, pages
  4904--4916. PMLR.

\bibitem[{Jurgens et~al.(2012)Jurgens, Mohammad, Turney, and
  Holyoak}]{jurgens-etal-2012-semeval}
David Jurgens, Saif Mohammad, Peter Turney, and Keith Holyoak. 2012.
\newblock \href {https://aclanthology.org/S12-1047} {{S}em{E}val-2012 task 2:
  Measuring degrees of relational similarity}.
\newblock In \emph{*{SEM} 2012: The First Joint Conference on Lexical and
  Computational Semantics {--} Volume 1: Proceedings of the main conference and
  the shared task, and Volume 2: Proceedings of the Sixth International
  Workshop on Semantic Evaluation ({S}em{E}val 2012)}, pages 356--364,
  Montr{\'e}al, Canada. Association for Computational Linguistics.

\bibitem[{Kolluru(2019)}]{Kolluru2019ANA}
Sethu~Hareesh Kolluru. 2019.
\newblock A neural architecture to learn image-text joint embedding.
\newblock In \emph{Stanford}.

\bibitem[{Krishna et~al.(2017)Krishna, Zhu, Groth, Johnson, Hata, Kravitz,
  Chen, Kalantidis, Li, Shamma, Bernstein, and Fei-Fei}]{krishnavisualgenome}
Ranjay Krishna, Yuke Zhu, Oliver Groth, Justin Johnson, Kenji Hata, Joshua
  Kravitz, Stephanie Chen, Yannis Kalantidis, Li-Jia Li, David~A. Shamma,
  Michael~S. Bernstein, and Li~Fei-Fei. 2017.
\newblock \href {https://doi.org/10.1007/s11263-016-0981-7} {Visual genome:
  Connecting language and vision using crowdsourced dense image annotations}.
\newblock \emph{International Journal of Computer Vision}, 123.

\bibitem[{Levy and Goldberg(2014)}]{levy2014c}
Omer Levy and Yoav Goldberg. 2014.
\newblock \href
  {https://proceedings.neurips.cc/paper/2014/file/feab05aa91085b7a8012516bc3533958-Paper.pdf}
  {{Neural Word Embedding as Implicit Matrix Factorization}}.
\newblock In \emph{Advances in Neural Information Processing Systems},
  volume~27. Curran Associates, Inc.

\bibitem[{Levy et~al.(2015)Levy, Goldberg, and Dagan}]{levy2015improving}
Omer Levy, Yoav Goldberg, and Ido Dagan. 2015.
\newblock Improving distributional similarity with lessons learned from word
  embeddings.
\newblock \emph{Transactions of the association for computational linguistics},
  3:211--225.

\bibitem[{Li et~al.(2022{\natexlab{a}})Li, Andreeto, Ranzato, and
  Perona}]{caltech101}
Fei-Fei Li, Marco Andreeto, Marc'Aurelio Ranzato, and Pietro Perona.
  2022{\natexlab{a}}.
\newblock \href {https://doi.org/10.22002/D1.20086} {Caltech 101}.

\bibitem[{Li et~al.(2022{\natexlab{b}})Li, Li, Xiong, and Hoi}]{li2022blip}
Junnan Li, Dongxu Li, Caiming Xiong, and Steven Hoi. 2022{\natexlab{b}}.
\newblock Blip: Bootstrapping language-image pre-training for unified
  vision-language understanding and generation.
\newblock In \emph{International Conference on Machine Learning}, pages
  12888--12900. PMLR.

\bibitem[{Li et~al.(2021)Li, Selvaraju, Gotmare, Joty, Xiong, and Hoi}]{albef}
Junnan Li, Ramprasaath~R. Selvaraju, Akhilesh~Deepak Gotmare, Shafiq Joty,
  Caiming Xiong, and Steven Hoi. 2021.
\newblock \href {https://doi.org/10.48550/ARXIV.2107.07651} {Align before fuse:
  Vision and language representation learning with momentum distillation}.

\bibitem[{Li et~al.(2020)Li, Yin, Li, Zhang, Hu, Zhang, Wang, Hu, Dong, Wei
  et~al.}]{li2020oscar}
Xiujun Li, Xi~Yin, Chunyuan Li, Pengchuan Zhang, Xiaowei Hu, Lei Zhang, Lijuan
  Wang, Houdong Hu, Li~Dong, Furu Wei, et~al. 2020.
\newblock Oscar: Object-semantics aligned pre-training for vision-language
  tasks.
\newblock In \emph{European Conference on Computer Vision}, pages 121--137.
  Springer.

\bibitem[{Lin et~al.(2014)Lin, Maire, Belongie, Hays, Perona, Ramanan,
  Doll{\'a}r, and Zitnick}]{lin2014microsoft}
Tsung-Yi Lin, Michael Maire, Serge Belongie, James Hays, Pietro Perona, Deva
  Ramanan, Piotr Doll{\'a}r, and C~Lawrence Zitnick. 2014.
\newblock Microsoft coco: Common objects in context.
\newblock In \emph{European conference on computer vision}, pages 740--755.
  Springer.

\bibitem[{Lu et~al.(2018)Lu, Whitehead, Huang, Ji, and Chang}]{lu2018entity}
Di~Lu, Spencer Whitehead, Lifu Huang, Heng Ji, and Shih-Fu Chang. 2018.
\newblock Entity-aware image caption generation.
\newblock \emph{arXiv preprint arXiv:1804.07889}.

\bibitem[{Luo et~al.(2018)Luo, Price, Cohen, and
  Shakhnarovich}]{luo2018discriminability}
Ruotian Luo, Brian Price, Scott Cohen, and Gregory Shakhnarovich. 2018.
\newblock Discriminability objective for training descriptive captions.
\newblock \emph{arXiv preprint arXiv:1803.04376}.

\bibitem[{Mikolov et~al.(2013{\natexlab{a}})Mikolov, Chen, Corrado, and
  Dean}]{Mikolov2013EfficientEO}
Tomas Mikolov, Kai Chen, Gregory~S. Corrado, and Jeffrey Dean.
  2013{\natexlab{a}}.
\newblock Efficient estimation of word representations in vector space.
\newblock In \emph{ICLR}.

\bibitem[{Mikolov et~al.(2013{\natexlab{b}})Mikolov, Yih, and
  Zweig}]{mikolov-etal-2013-linguistic}
Tomas Mikolov, Wen-tau Yih, and Geoffrey Zweig. 2013{\natexlab{b}}.
\newblock \href {https://aclanthology.org/N13-1090} {Linguistic regularities in
  continuous space word representations}.
\newblock In \emph{Proceedings of the 2013 Conference of the North {A}merican
  Chapter of the Association for Computational Linguistics: Human Language
  Technologies}, pages 746--751, Atlanta, Georgia. Association for
  Computational Linguistics.

\bibitem[{Oord et~al.(2018)Oord, Li, and Vinyals}]{oord2018representation}
Aaron van~den Oord, Yazhe Li, and Oriol Vinyals. 2018.
\newblock Representation learning with contrastive predictive coding.
\newblock \emph{arXiv preprint arXiv:1807.03748}.

\bibitem[{Pennington et~al.(2014)Pennington, Socher, and
  Manning}]{pennington-etal-2014-glove}
Jeffrey Pennington, Richard Socher, and Christopher Manning. 2014.
\newblock \href {https://doi.org/10.3115/v1/D14-1162} {{G}lo{V}e: Global
  vectors for word representation}.
\newblock In \emph{Proceedings of the 2014 Conference on Empirical Methods in
  Natural Language Processing ({EMNLP})}, pages 1532--1543, Doha, Qatar.
  Association for Computational Linguistics.

\bibitem[{Pilehvar et~al.(2018)Pilehvar, Kartsaklis, Prokhorov, and
  Collier}]{pilehvar-etal-2018-card}
Mohammad~Taher Pilehvar, Dimitri Kartsaklis, Victor Prokhorov, and Nigel
  Collier. 2018.
\newblock \href {https://doi.org/10.18653/v1/D18-1169} {Card-660: {C}ambridge
  rare word dataset - a reliable benchmark for infrequent word representation
  models}.
\newblock In \emph{Proceedings of the 2018 Conference on Empirical Methods in
  Natural Language Processing}, pages 1391--1401, Brussels, Belgium.
  Association for Computational Linguistics.

\bibitem[{Radford et~al.(2021)Radford, Kim, Hallacy, Ramesh, Goh, Agarwal,
  Sastry, Askell, Mishkin, Clark et~al.}]{radford2021learning}
Alec Radford, Jong~Wook Kim, Chris Hallacy, Aditya Ramesh, Gabriel Goh,
  Sandhini Agarwal, Girish Sastry, Amanda Askell, Pamela Mishkin, Jack Clark,
  et~al. 2021.
\newblock Learning transferable visual models from natural language
  supervision.
\newblock \emph{arXiv preprint arXiv:2103.00020}.

\bibitem[{Rubenstein et~al.(1965)Rubenstein, Goodenough, and
  Goodenough}]{10.1145/365628.365657}
Herbert Rubenstein, John~B. Goodenough, and John~B. Goodenough. 1965.
\newblock \href {https://doi.org/10.1145/365628.365657} {Contextual correlates
  of synonymy}.
\newblock \emph{Commun. ACM}, 8(10):627–633.

\bibitem[{Schroff et~al.(2015)Schroff, Kalenichenko, and
  Philbin}]{schroff2015facenet}
Florian Schroff, Dmitry Kalenichenko, and James Philbin. 2015.
\newblock Facenet: A unified embedding for face recognition and clustering.
\newblock In \emph{Proceedings of the IEEE conference on computer vision and
  pattern recognition}, pages 815--823.

\bibitem[{Sharma et~al.(2018)Sharma, Ding, Goodman, and
  Soricut}]{sharma-etal-2018-conceptual}
Piyush Sharma, Nan Ding, Sebastian Goodman, and Radu Soricut. 2018.
\newblock \href {https://doi.org/10.18653/v1/P18-1238} {Conceptual captions: A
  cleaned, hypernymed, image alt-text dataset for automatic image captioning}.
\newblock In \emph{Proceedings of the 56th Annual Meeting of the Association
  for Computational Linguistics (Volume 1: Long Papers)}, pages 2556--2565,
  Melbourne, Australia. Association for Computational Linguistics.

\bibitem[{Simonyan and Zisserman(2014)}]{simonyan2014very}
Karen Simonyan and Andrew Zisserman. 2014.
\newblock Very deep convolutional networks for large-scale image recognition.
\newblock \emph{arXiv preprint arXiv:1409.1556}.

\bibitem[{Tan and Bansal(2019)}]{tan2019lxmert}
Hao Tan and Mohit Bansal. 2019.
\newblock Lxmert: Learning cross-modality encoder representations from
  transformers.
\newblock \emph{arXiv preprint arXiv:1908.07490}.

\bibitem[{Thompson(2000)}]{thompson2000canonical}
Bruce Thompson. 2000.
\newblock Canonical correlation analysis.

\bibitem[{Wang et~al.(2022)}]{cvprtut}
Jianfeng Wang et~al. 2022.
\newblock Recent advances in vision-and-language pre-training (tutorial).

\bibitem[{Young et~al.(2014)Young, Lai, Hodosh, and
  Hockenmaier}]{young2014image}
Peter Young, Alice Lai, Micah Hodosh, and Julia Hockenmaier. 2014.
\newblock From image descriptions to visual denotations: New similarity metrics
  for semantic inference over event descriptions.
\newblock \emph{Transactions of the Association for Computational Linguistics},
  2:67--78.

\bibitem[{Zeng et~al.(2021)Zeng, Zhang, and Li}]{xvlm}
Yan Zeng, Xinsong Zhang, and Hang Li. 2021.
\newblock \href {https://doi.org/10.48550/ARXIV.2111.08276} {Multi-grained
  vision language pre-training: Aligning texts with visual concepts}.

\bibitem[{Zhang et~al.(2020)Zhang, Kishore, Wu, Weinberger, and
  Artzi}]{zhang2020bertscore}
Tianyi Zhang, Varsha Kishore, Felix Wu, Kilian~Q. Weinberger, and Yoav Artzi.
  2020.
\newblock \href {http://arxiv.org/abs/1904.09675} {Bertscore: Evaluating text
  generation with bert}.

\bibitem[{Zhou et~al.(2020)Zhou, Palangi, Zhang, Hu, Corso, and
  Gao}]{zhou2020unified}
Luowei Zhou, Hamid Palangi, Lei Zhang, Houdong Hu, Jason Corso, and Jianfeng
  Gao. 2020.
\newblock Unified vision-language pre-training for image captioning and vqa.
\newblock In \emph{Proceedings of the AAAI Conference on Artificial
  Intelligence}, volume~34, pages 13041--13049.

\bibitem[{Zhou et~al.(2022)Zhou, Girdhar, Joulin, Kr{\"a}henb{\"u}hl, and
  Misra}]{zhou2022detic}
Xingyi Zhou, Rohit Girdhar, Armand Joulin, Phillip Kr{\"a}henb{\"u}hl, and
  Ishan Misra. 2022.
\newblock Detecting twenty-thousand classes using image-level supervision.
\newblock \emph{arXiv preprint arXiv:2201.02605}.

\end{thebibliography}
